\RequirePackage{fix-cm}
\documentclass[twocolumn]{svjour3}          
 \pdfoutput=1
\smartqed  
\usepackage{graphicx}
\usepackage{amsmath}
\usepackage{mathrsfs}
\usepackage{amssymb}
\usepackage{url}
\usepackage{caption}
\usepackage{enumitem}
\usepackage{mathrsfs}
\usepackage{algorithm}
\usepackage{algpseudocode}
\usepackage{dsfont}
\usepackage{multirow}
\usepackage{booktabs}
\usepackage{wrapfig}
\usepackage{ulem}
\usepackage{array}
\usepackage{epstopdf}
\usepackage{times}
\usepackage{colortbl}
\usepackage[pagebackref=false,breaklinks=true,letterpaper=true,colorlinks,urlcolor=magenta,citecolor=blue,linkcolor=blue,bookmarks=false]{hyperref}
\def\eg{{\it{e.g.}}}
\def\etal{{\it{et al.}}}
\def\ie{{\it{i.e.}}}
\newcommand{\kk}[1]{\textcolor[rgb]{0.0,.0,.0}{#1}}
\newcommand{\kkk}[1]{\textcolor[rgb]{0.0,.0,.0}{#1}}

\newcommand{\tx}[1]{\textcolor[rgb]{0,.0,.0}{#1}}
\newcommand{\txx}[1]{\textcolor[rgb]{0,.0,.0}{#1}}

\usepackage{algorithmicx}
\floatname{algorithm}{Procedure}

\begin{document}
\begin{sloppypar}
\title{Learning to Detect Instance-level Salient Objects using Complementary Image Labels 
}


\author{Xin Tian* \and
        Ke Xu*    \and
        Xin Yang$\dagger$  \and
        Baocai Yin \and
        Rynson W.H. Lau$\dagger$ 
}


\institute{* Joint first authors, $\dagger$ joint corresponding authors. Rynson Lau leads this project.\\ \\
           Xin Tian* \at
           Dalian University of Technology and City University of Hong Kong.
           \and
           Ke Xu* \at
           City University of Hong Kong, Hong Kong SAR, China.
           \and
           Xin Yang$\dagger$ \at
           Dalian University of Technology, Dalian, China.
           \and
           Baocai Yin \at
           Dalian University of Technology and Pengcheng Lab, China.
           \and
           Rynson W.H. Lau$\dagger$ \at
           City University of Hong Kong, Hong Kong SAR, China.
}

\date{Received: date / Accepted: date}

\maketitle

\begin{abstract}
Existing salient instance detection (SID) methods typically learn from pixel-level annotated datasets. In this paper, we present the first weakly-supervised approach to the SID problem. Although weak supervision has been considered in general saliency detection, it is mainly based on using class labels for object localization. However, it is non-trivial to use only class labels to learn instance-aware saliency information, as salient instances with high semantic affinities may not be easily separated by the labels.
As the subitizing information provides an instant judgement on the number of salient items, it is naturally related to detecting salient instances and may help separate instances of the same class while grouping different parts of the same instance.
Inspired by this observation, we propose to use class and subitizing labels as weak supervision for the SID problem. We propose a novel weakly-supervised network with three branches:
a Saliency Detection Branch leveraging class consistency information to locate candidate objects;
a Boundary Detection Branch exploiting class discrepancy information to delineate object boundaries;
and a Centroid Detection Branch using subitizing information to detect salient instance centroids.
This complementary information is then fused to produce a salient instance map.
\kk{To facilitate the learning process, we further propose a progressive training scheme to reduce label noise and the corresponding noise learned by the model, via reciprocating the model with progressive salient instance prediction and model refreshing.}
Our extensive evaluations show that the proposed method plays favorably against carefully designed baseline methods adapted from related tasks.
\end{abstract}

\begin{figure}[tb]
\begin{center}
\includegraphics[width=0.99\linewidth]{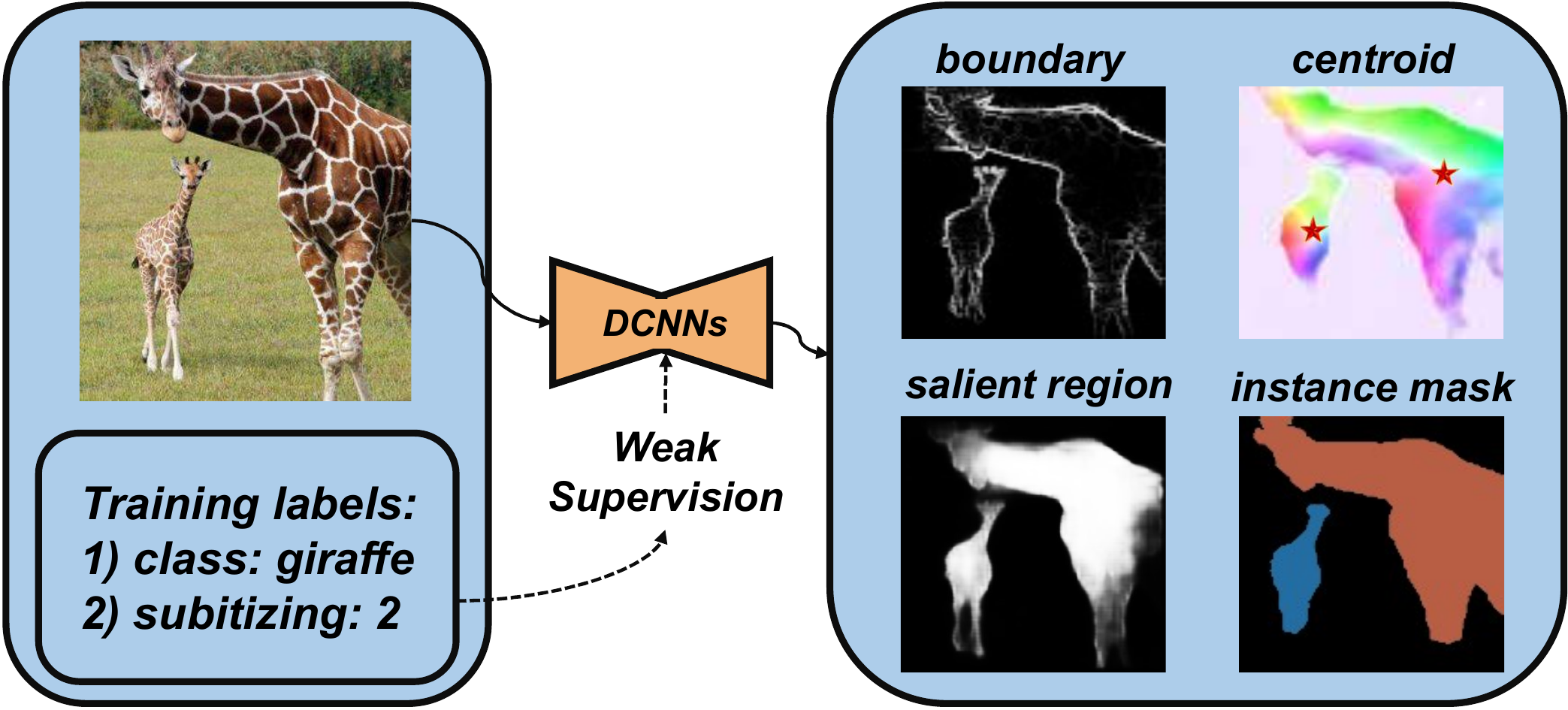}
\end{center}\vspace{-3mm}
\caption{Our key idea is to leverage complementary image-level labels (class and subitizing) to train a salient instance detection model in a weakly-supervised manner, via synergically learning to predict salient objects, detecting object boundaries and locating instance centroids.}
\label{fig:teaser}
\vspace{-3mm}
\end{figure}

\section{Introduction}
\label{sec:intro}

Salient Object Detection (SOD) is a long-standing vision task that aims to segment visually salient objects in a scene. It often serves as a core step for downstream vision tasks like video object segmentation~\cite{wang2015saliency}, object proposal generation~\cite{alexe2012measuring}, and image cropping~\cite{wang2017deep}. Recent deep learning-based SOD methods have achieved a significant performance progress~\cite{wang2017learning,zhuge2018boundary,xu2019structured,su2019selectivity,zhao2019pyramid,hou2017deeply,wang2018detect}, benefited from the powerful representation learning capability of neural networks and large-scale pixel-level annotated training data.
Since annotating pixel-level labels is extremely tedious, there are some works~\cite{wang2017learning,zeng2019multi} that aim to explore cheaper image-level labels (\eg, class labels) to train SOD models in a weakly-supervised manner.

Salient Instance Detection (SID) goes further from SOD as it aims to differentiate individual salient instances. {This instance-level saliency information can benefit vision tasks that require fine-grained scene understanding, \eg, \txx{object rank~\cite{siris2020inferring}}, image captioning~\cite{karpathy2015deep}, image editing~\cite{NEURIPS2018_653ac11c} and semantic segmentation~\cite{xin2021ntsp}. However, existing SID methods \cite{fan2019s4net,li2017instance,zhang2016unconstrained} still rely on large-scale annotated ground truth masks in order to learn how to segment salient instances with their boundaries delineated.}
Hence, it is worthwhile studying the SID problem from the weakly-supervised perceptive by using cheaper image-level labels.

A straightforward solution to the weakly-supervised SID problem is to use class labels for training, like the weakly-supervised SOD methods~\cite{wang2017learning,zeng2019multi}. However, using just class labels to learn a SID model is non-trivial for two reasons. First, while class labels can help detect semantically predominant regions~\cite{zhou2016learning}, there is no guarantee that the detected regions are visually salient. Second, objects of the same class may not be \kk{easily} distinguished due to their high semantic affinity.
We observe that subitizing refers to the number of certain objects and is therefore naturally related to saliency instance detection. By predicting the number of salient objects, we may use it as a global supervision to help separate instances of the same class while clustering different parts of an instance with diverse appearances into one.
Inspired by this insight, we propose to learn a weakly-supervised SID network (denoted as WSID-Net) using class and subitizing labels.

Our WSID-Net consists of three synergic branches:
a salient object detection branch is proposed to locate candidate salient objects while a boundary detection branch is proposed to delineate their boundaries, both by exploiting semantics from the class labels; and a centroid detection branch is proposed to detect the centroid of each salient instance, by leveraging saliency cues from the subitizing labels. This information is fused to obtain the salient instance map.
\kk{To facilitate the learning process, we propose a Progressive Training Scheme (PTS) to reduce the noise generated in our salient object detection branch (\eg, incomplete object proposals and cluttered background objects), by reciprocally updating the branch using generated pseudo labels and refreshing the branch in a self-supervised manner.}
To demonstrate the effectiveness of the proposed model, we compare it with a variety of baselines adapted from related tasks on the standard benchmark~\cite{li2017instance}.

To summarize, this work has four main contributions:
\begin{itemize}
\item To the best of our knowledge, we propose the first weakly-supervised method for salient instance detection, \kk{which only requires image-level class and subitizing labels to obtain salient instance maps}.
\item We propose a novel \kk{network (WSID-Net)}, with a novel centroid-based subitizing loss to exploit salient instance number, a novel Boundary Enhancement module to learn  instance boundaries, and a novel Cross-layer Attention module to enhance cross-layer context feature learning of centroids and the boundaries.
\item We propose a \kk{novel} Progressive Training Scheme, \kk{to facilitate the learning of the saliency detection branch by reducing the noise in a self-supervised manner}.
\item We conduct extensive experiments to analyze the proposed method, and verify its superiority against baselines adapted from related state-of-the-art approaches.
\end{itemize}


\section{Related Work}

\subsection{Salient Instance Detection}
Existing SID methods are fully-supervised.
Zhang~\etal~\cite{zhang2016unconstrained} propose to detect salient instances with bounding boxes, and propose a {MAP}-based optimization framework to regress a large amount of pre-defined bounding boxes into a compact number of instance-level bounding boxes of high confidences.
However, this method based on bounding boxes cannot detect salient instances with accurately delineated boundaries.
Other works predict pixel-wise masks for the detected salient instances, and typically rely on large amount of manually annotated ground truth labels.
Specifically, Li~\etal~\cite{li2017instance} propose to first predict the saliency mask and instance-aware saliency contour, and then apply the Multi-scale Combinatorial Grouping (MCG) algorithm~\cite{APBMM2014} to extract instance-level masks.
Fan~\etal~\cite{fan2019s4net} propose an end-to-end SID network based on the object detection model FPN~\cite{lin2017feature}, with a segmentation branch to segment the salient instances.

Unlike these existing SID methods, we propose in this paper to train a weakly-supervised network, which only requires image-level class and subitizing labels.

\kkk{The work presented in this paper extends our BMVC oral paper~\cite{tian2020weakly} in three aspects.
First, we provide a more comprehensive literature survey on the weakly supervised salient instance detection task and other relevant works.
Second, we note that the earlier method~\cite{tian2020weakly} typically suffers from the salient instance incompleteness problem, due to the noise generated in both salient object detection and boundary detection branches.
To address this problem, we propose a Cross-layer Attention module here to learn boundary and centroid features, and a self-supervised Progressive Training Scheme to reduce the noise in the salient object detection branch.
Third, we perform more experiments to analyze the properties of our method and show its effectiveness over existing state-of-the-art approaches.
}

\subsection{Salient Object Detection}
SOD methods aim at detecting salient objects in a scene without differentiating the detected instances.
Liu~\etal~\cite{liu2010learning} formulate the SOD task as a binary segmentation problem for segmenting out the visually conspicuous objects of an image via color and contrast histogram based priors.

Traditional methods propose to leverage different hand-crafted priors to detect salient objects, \eg, image colors and luminance~\cite{achanta2009frequency}, global and local contrast priors~\cite{perazzi2012saliency,cheng2014global}, and {background geometric distance prior}~\cite{yang2013saliency}.
Recently, deep learning based SOD methods achieve superior performances on the standard SOD {benchmarks}~\cite{yang2013saliency,li2014secrets,shi2015hierarchical,jiang2013salient,wang2017learning,cheng2014global}. 
\txx{Among them, several methods explore boundary information for salient object detection.
Xu~\etal~\cite{xu2019structured} propose a CRF-based architecture to refine boundaries of both deep features and saliency maps in a coarse-to-fine manner. 
Some methods~\cite{zhuge2018boundary,zhou2020interactive,wei2020label,su2019selectivity} propose to formulate saliency and edge detection with two network branches as multi-task learning.
%
%
Feature fusion strategies have also been widely explored in salient object detection.
DSS~\cite{hou2017deeply}, DGRL~\cite{wang2018detect}, and MINet~\cite{pang2020multi} integrate multi-level features in the top-down direction. 
In GBMPM~\cite{zhang2018bi} and PAGE-Net~\cite{wang2019salient}, multi-level saliency features are fused in both top-down and bottom-up directions, to detect salient objects of varying scales.
F3Net~\cite{wei2020f3net} and PFPN~\cite{wang2020progressive} propose to fuse features progressively, to enrich saliency features with recurrent feedback information. 
Attention mechanism has also been exploited to reweigh multi-scale features in order to suppress noise and enhance context learning, via dynamic weight decay scheme~\cite{gao2020highly}, mutual relation learning of object parts~\cite{chen2020learning}, gate-based interference control~\cite{zhao2020suppress}, and spatial-/channel-wise attentions on different features~\cite{zhao2019pyramid}.
}
%
In particular, He \etal \cite{he2017delving} propose to leverage numerical representation of subitizing to enrich spatial representations of salient objects.
These methods are typically benefited from the powerful learning ability of deep neural networks as well as large-scale annotated ground truth data.

To alleviate the data annotation efforts, \kk{many weakly-supervised SOD methods are proposed, by investigating different approaches of generating pseudo saliency labels.}
A method leverage subitizing information alone for refining saliency prediction 
Some methods~\cite{zhang2017supervision,zhang2018deep,liu2021weakly} propose to use traditional \kk{SOD} methods to generate pseudo labels for \kk{training deep saliency models. Some other methods}~\cite{wang2017learning,zeng2019multi} propose to train weakly-supervised deep models using object class labels and class activation maps (CAMs)~\cite{zhou2016learning}.
\kk{There are also some methods \cite{li2018contour,zhang2020weakly}} that propose to combine pre-trained contour networks with segment proposals~\cite{li2018contour} or scribbles~\cite{zhang2020weakly} to generate pseudo labels for training saliency detection networks.

However, existing weakly-supervised SOD methods cannot be directly applied to our problem, as class labels along do not provide instance-level information. In this paper, we propose to use class and subitizing labels to train our SID model.

\subsection{Noise Reduction}

Noise commonly exists in \kk{a weakly-supervised setting, typically when the task is a pixel-level prediction and the supervision is provided at the image-level. Existing methods typically rely on auxiliary full-annotated labels (referred to as clean labels) or pre-trained models to regularize the noise.
\kk{Hu~\etal~\cite{hu2019weakly} formulate it as a multi-task learning problem, in which the networks trained on a small set of clean labels can help regularize the noise in the networks trained on a large set of weak labels.}
\txx{Zhang~\etal~\cite{zhang2020learning} propose a noise-aware method for learning a disentangled clean saliency detector from noisy labels.}
\kk{Lu~\etal~\cite{lu2016learning} propose a sparse learning model to learn the noise statistics from over-segmented superpixels, while Zhu~\etal~\cite{zhu2019learning} propose to filter out noisy segment proposals with low matching scores. Both methods rely on additional pre-trained models for noise reduction.}
%
}

Unlike the above methods, we do not leverage clean labels or pre-trained proposals as assistances. \kk{We achieve this goal by training our salient object detection branch in a progressive manner,} via model refreshing and pseudo label regeneration.

\section{Methodology}

\begin{figure*}[!t]
\begin{center}
\includegraphics[width=0.99\linewidth]{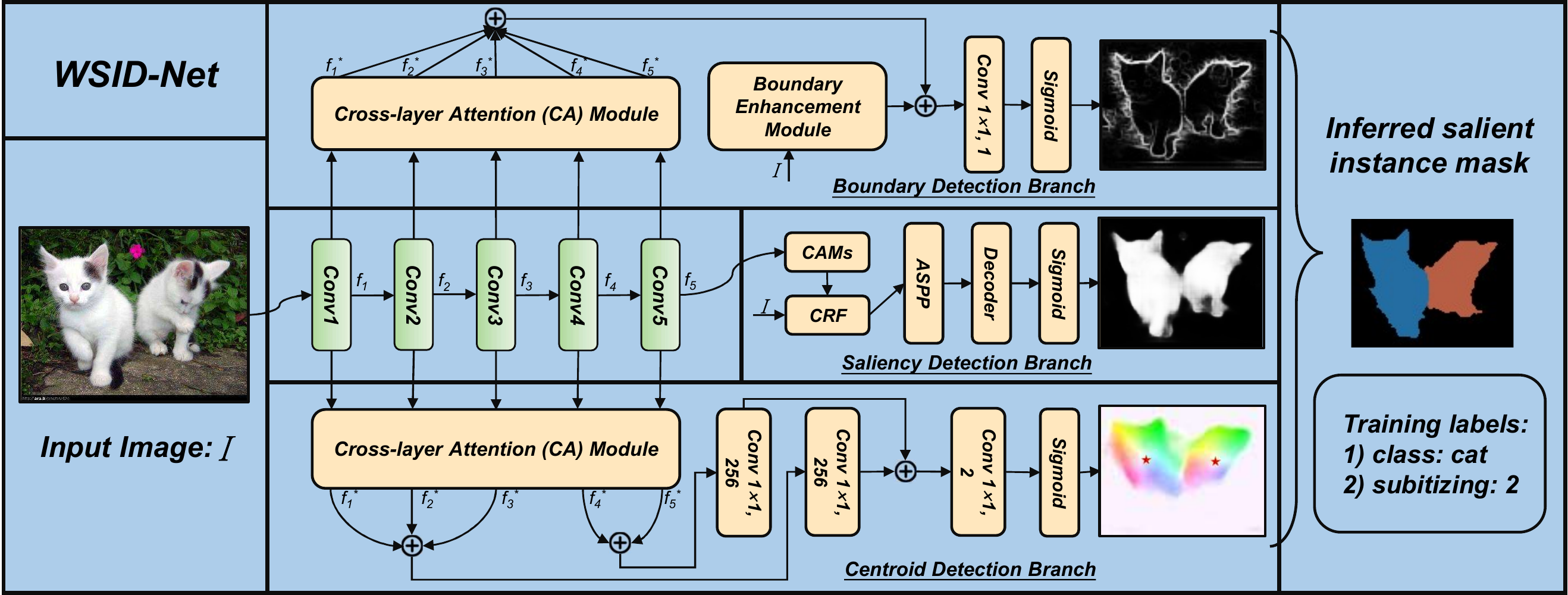}
\end{center}\vspace{-3mm}
\caption{\tx{Pipeline overview}. Our SID model is trained using only image-level class and subitizing labels. It has three synergic branches: (1) a Boundary Detection Branch for detecting object boundaries using class discrepancy information; (2) a Saliency Detection Branch for detecting objects using class consistency information; and (3) a Centroid Detection Branch for detecting salient instance centroids using subitizing information. A random walk method is further applied to fuse these information to obtain a final salient instance mask.}
\label{fig:pipeline}
\vspace{-3mm}
\end{figure*}

Class labels are widely explored in weakly-supervised SOD methods for learning to localize candidate objects, based on the pixel-level semantic affinity derived from the network responses to the class labels. However, class labels lack instance-level information, causing over- and under-detection when salient instances are from the same category.
We note that subitizing, which is a cheap image-level label that denotes the number of salient instances of a scene, can serve as a complementary supervision to the class labels to provide instance-related information.
Hence, we propose to use both class and subitizing labels to address our weakly-supervised SID problem. To this end, we propose a weakly-supervised SID network (WSID-Net), as shown in Figure~\ref{fig:pipeline}.

The proposed WSID-Net has three branches:
\begin{enumerate}
\item A \textbf{\txx{$Saliency$} $Detection$ $Branch$} for locating candidate salient objects. This saliency detection branch is based on Deeplab~\cite{chen2017rethinking} by modifying its last layer for binary prediction. We propose a novel Progressive Training Scheme (PTS)  to self-correct the noise coming from the weak labels and the corresponding noise learned by this saliency branch.
\item A \textbf{$Centroid$ $Detection$ $Branch$} for detecting the centroids of salient instances, where subitizing knowledge is utilized in a novel loss function to provide regularization on the global number of instance centroids.
\item A \textbf{$Boundary$ $Detection$ $Branch$} for delineating salient instance boundaries, where a novel Boundary Enhancement (BE) module is introduced to resolve the discontinuity problem of detected boundaries.
\end{enumerate}
Finally, we propose a novel Cross-layer Attention (CA) module for the Centroid Detection Branch and Boundary Detection Branch to learn the context information for detecting centroids and boundaries, respectively.

\subsection{Centroid Detection Branch}\label{sec:322}

Detecting object centroids is crucial to separating object instances in a weakly-supervised scheme. Unlike existing semantic (instance) segmentation methods~\cite{jiwoon2019weakly,Neven2019InstanceSB,zhou2018weakly,cholakkal2019object,zhu2019learning,laradji2019masks} that detect the centroids based on network responses to the class labels, we propose to introduce subitizing information to explicitly supervise the salient centroid detection process.

\subsubsection{Centroid-based Subitizing Loss}\label{lsu_loss}

It has been shown that penalizing the centroid loss~\cite{jiwoon2019weakly,Neven2019InstanceSB} helps cluster local pixels with high semantic affinities. However, this typically fails when salient instances from the same object category have varying shapes and appearances.
The reason is that the clustering process of local pixels lacks global saliency supervision. Hence, we introduce the centroid-based subitizing loss $\mathcal{L}_{\mathcal{SU}}$ to resolve this problem. We use subitizing to explicitly supervise the number of predicted centroids\txx{, which is implicitly related to the learned offset vectors as $\mathcal{L}_{su}$ is back-propagated to the centroid detection branch during training, to guide the centroid-aware pixel clustering process. The detailed formulation is discussed below.}

\txx{The Centroid Detection Branch predicts an offset vector map $\mathcal{V}\in\mathbb{R}^{W \times H \times 2}$, where $W \times H$ denotes the spatial size of the map, and each 2D vector $v_{i} \in \mathcal{V}$ indicates the vertical and horizontal distances of the $i^{th}$ pixel from its associated instance centroid. 
We follow~\cite{jiwoon2019weakly} to iteratively derive $\mathcal{V}$ as:
\begin{equation}
    v_{i}^{m+1} = v_{i}^{m} + v_{v_i^{m}+p_i},
    \label{eq:1}
\end{equation}
where $p_i$ is the coordinates of the $i^{th}$ pixel, $m$ is the iteration number, and $v_i^m+p_i$ indexes the the current centroid that the offset of the $i^{th}$ pixel points to.
Ideally, Eq.~\ref{eq:1} would converge within a few iterations when $v_{i}^{m+1} = v_{i}^{m}$ and the offset of the centroid is zero, and yields a set of centroids that represent the instances.
Pixel $i$ can then be assigned to its corresponding centroid $c_n$ by measuring its distance from the centroid, as described by:
\begin{equation}
    c_{i\rightarrow n} = \mathop{\arg\min}_{n} \ \ \| v_{i} + p_i - p_{c_n} \|.
\end{equation}
We then use the saliency map $\mathcal{S}$ of the saliency detection branch to filter out non-salient instances by computing their $IoU=(\mathcal{SI}_{n} \cap \mathcal{S})/\mathcal{SI}_{n}>\theta$, to obtain a set of saliency instances $\mathcal{SI}^{*}=\{\mathcal{SI}_{1}, \mathcal{SI}_{2}, ..., \mathcal{SI}_{T^{*}}\}$, where $T^{*}$ represents the number of predicted salient instances.
Finally, we use MSE to measure $\mathcal{L}_{su}$ as: 
\begin{equation}
    \mathcal{L}_{su} = MSE(T^{*}, T), 
\end{equation}
where $T$ is the subitizing ground truth, and $T^{*}$ denotes the number of predicted centroids extracted from the offset vectors of the pixels in the salient region. 
Note that the loss $\mathcal{L}_{su}$ is back-propagated to update the offset vectors only in the salient region, which avoids the learning process of instance centroid detection being distracted by the non-salient background.
The gradient $\delta$ of $\mathcal{L}_{su}$ is calculated as:
\begin{equation}
\centering
    \delta = \frac{1}{K} \cdot \frac{\partial\mathcal{L}_{su}}{\partial\mathcal{V}^{*}},
\end{equation}
where $\mathcal{V}^{*}$ are the offset vectors in the salient region, and $K$ is the total number of offset vectors in $\mathcal{V}^{*}$.
}

\begin{figure}[tb]
\centering
\includegraphics[width=.13\textwidth]{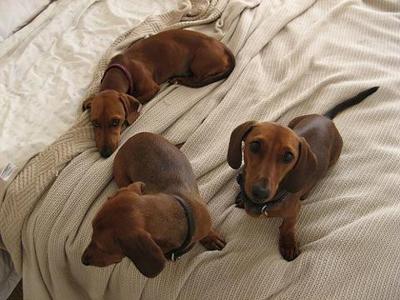}
\includegraphics[width=.13\textwidth]{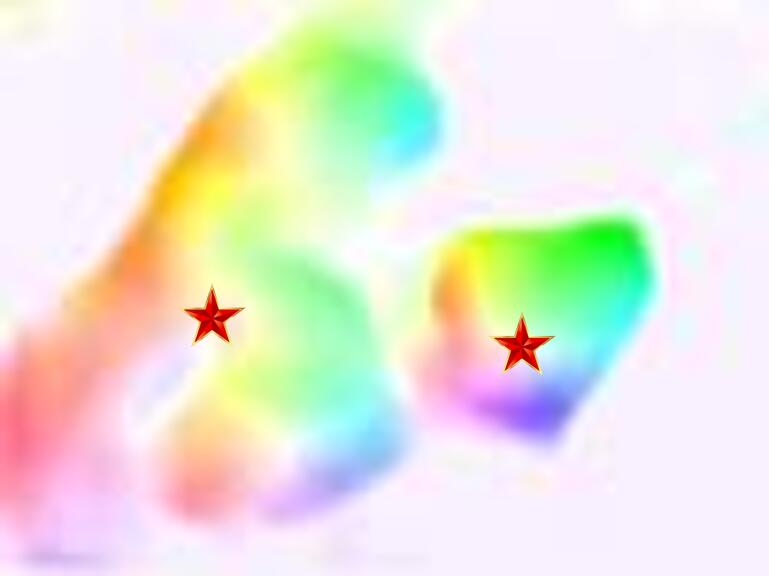}
\includegraphics[width=.13\textwidth]{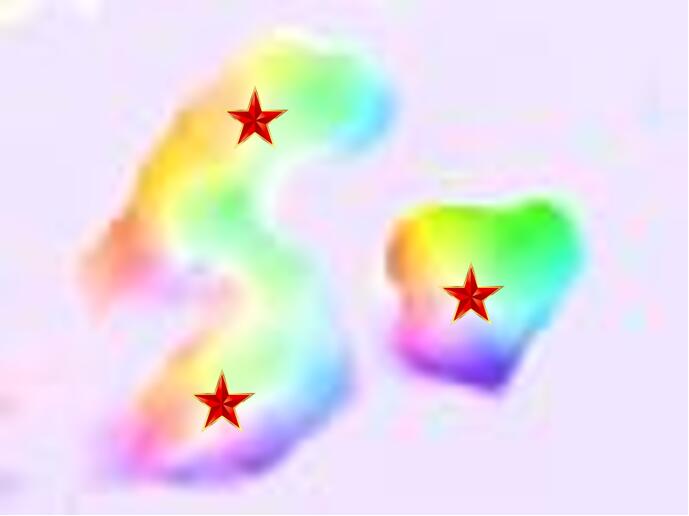}
\centering
\begin{minipage}[t]{0.13\textwidth}
\centering
\scriptsize{\textbf{(a) input\\image}}
\end{minipage}
\begin{minipage}[t]{0.13\textwidth}
\centering
\scriptsize{\textbf{(b) centroid\\(w/o $\mathcal{L}_{\mathcal{SU}}$)}}
\end{minipage}
\begin{minipage}[t]{0.13\textwidth}
\centering
\scriptsize{\textbf{(c) centroid\\(w/ $\mathcal{L}_{\mathcal{SU}}$)}}
\end{minipage}
\\
\includegraphics[width=.13\textwidth]{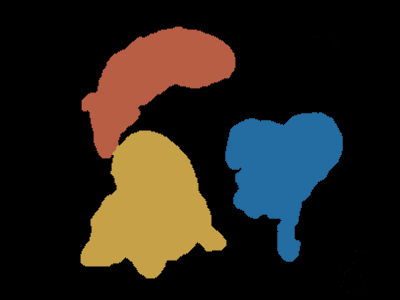}
\includegraphics[width=.13\textwidth]{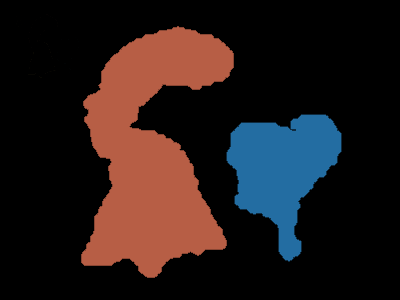}
\includegraphics[width=.13\textwidth]{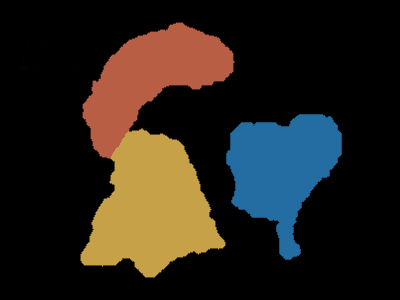}
\\

\begin{minipage}[t]{0.13\textwidth}
\centering
\scriptsize{\textbf{(d) ground \\truth}}
\end{minipage}
\begin{minipage}[t]{0.13\textwidth}
\centering
\scriptsize{\textbf{(e) instance mask\\(w/o $\mathcal{L}_{\mathcal{SU}}$)}}
\end{minipage}
\begin{minipage}[t]{0.13\textwidth}
\centering
\scriptsize{\textbf{(f) instance mask\\(w/ $\mathcal{L}_{\mathcal{SU}}$)}}
\end{minipage}
\vspace{-3mm}
\caption{Visualization of the centroid detection branch with and without $\mathcal{L}_{\mathcal{SU}}$. \kk{Using class labels alone fails to train the network to detect instance centroids (denoted by red stars) if they have similar appearances (b), resulting in wrong segmentations (e). In contrast, our proposed subitizing loss can segment these salient objects in instance-level (f), by learning to identify the correct number of salient instances (c).} 
}
\label{fig:wsu}
\vspace{-3mm}
\end{figure}

Figure~\ref{fig:wsu} visualizes the results from centroid detection and the corresponding instance segmentation, with and without using the centroid-based subitizing $\mathcal{L}_{\mathcal{SU}}$ loss function. We can see that the network groups the two dogs into one when not using $\mathcal{L}_{\mathcal{SU}}$, as these two dogs have similar appearances and lie next to each other (Figure~\ref{fig:wsu}(b,e)). By introducing $\mathcal{L}_{\mathcal{SU}}$, the network is able to predict a correct number of centroids, and generate reasonable salient instance masks compared with the ground truth (Figure~\ref{fig:wsu}(c,f)).

\subsubsection{Network Structure}
We adopt the image-to-image translation scheme, where our network outputs a 2D centroid map, in which the value of each pixel location indicates the offset vector to its instance centroid. The bottom part of Figure~\ref{fig:pipeline} shows the network structure of our centroid detection branch. Given an input image, we first extract multi-scale backbone features $f_{1}$ to $f_{5}$ and feed them to the \kk{Cross-layer Attention (CA) modules with boundary-aware features for joint refinement (to be discussed in Section~\ref{sec:da}).} We then fuse the high-level features to obtain \txx{$f_{h}$: $f_{h} = Conv(Concat({f_{4}}^{*},{f_{5}}^{*}))$}, which is further fused with the low-level features to produce the centroid map $\mathcal{V}$: \txx{$\mathcal{V} = \sigma(Conv(Conv(Concat(f_{h}, {f_{1}}^{*}, {f_{2}}^{*}, {f_{3}}^{*}))))$}.

\subsection{Boundary Detection Branch}\label{sec:321}

Boundaries provide strong cues for separating salient instances. Unlike fully-supervised SID methods that learn boundary-aware information based on pixel-level ground truth masks, we propose the Boundary Enhancement module to leverage the Canny prior~\cite{john1986} to delineate continuous instance boundaries.

\subsubsection{Boundary Enhancement (BE) Module}
\begin{figure}[h]
\begin{center}
\includegraphics[width=0.99\linewidth]{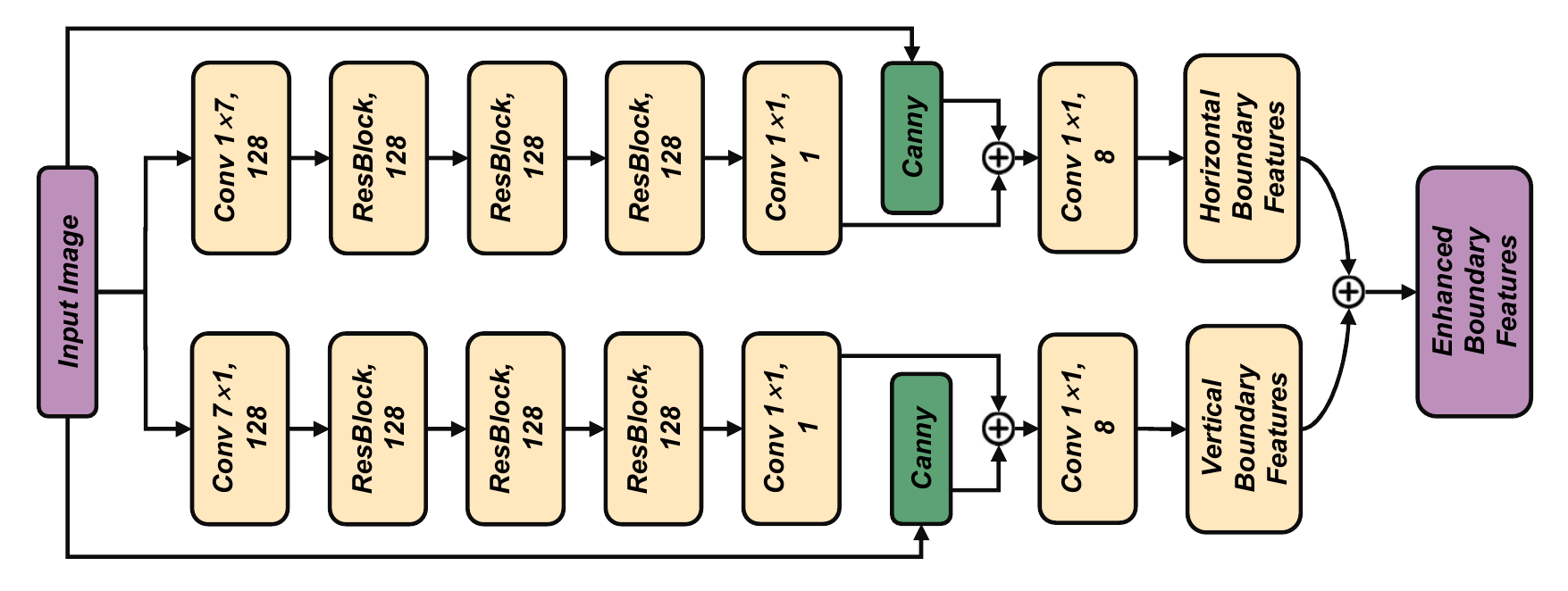}
\end{center}
\vspace{-3mm}
\centering
\caption{Boundary Enhancement (BE) module.}
\label{fig:BE}
\vspace{-3mm}
\end{figure}

We apply a random walk algorithm to search a salient instance from a centroid to its boundary. However, it may fail when part of the boundary is discontinuous as the random walk algorithm will also search the region outside the boundary. Hence, we propose the BE module to incorporate the edge prior for learning continuous instance boundaries, as shown in Figure~\ref{fig:BE}. Specifically, we first extract low-level features along the horizontal and vertical directions from the input image, by two $1\times7$ and $7\times1$ convolution layers. These low-level features are then fed into three Residual Blocks~\cite{he2016deep} for feature refinement, which are further concatenated with enriched edges computed from the Canny operator~\cite{john1986}. To compute the final enriched boundary features, another 1$\times$1 convolution layer is applied.

\begin{figure}[h]
\centering
\includegraphics[width=.09\textwidth]{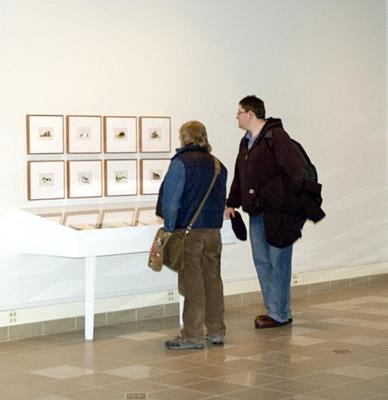}
\includegraphics[width=.09\textwidth]{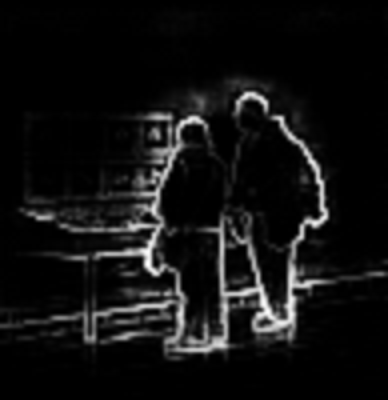}
\includegraphics[width=.09\textwidth]{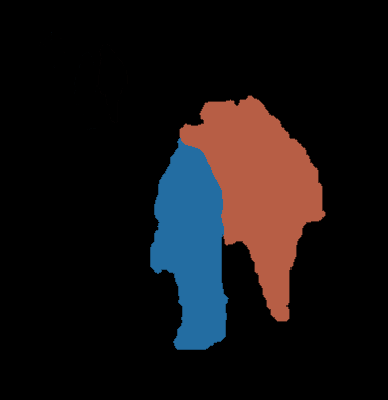}
\includegraphics[width=.09\textwidth]{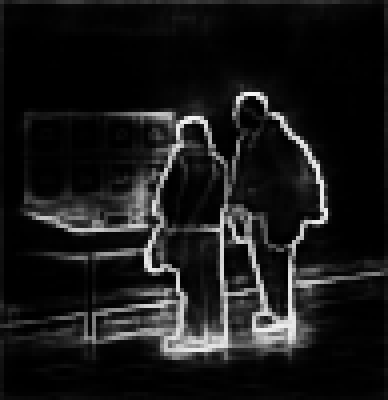}
\includegraphics[width=.09\textwidth]{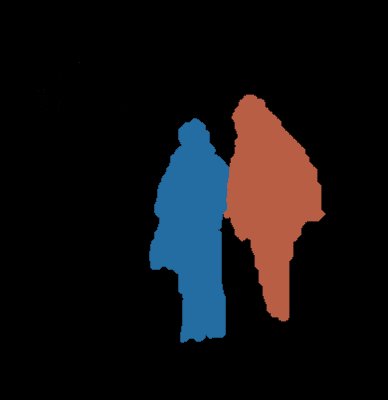}
\\
\vspace{0.1cm}
\centering
\includegraphics[width=.09\textwidth]{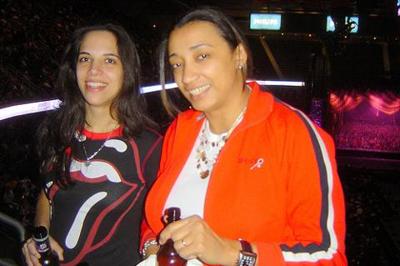}
\includegraphics[width=.09\textwidth]{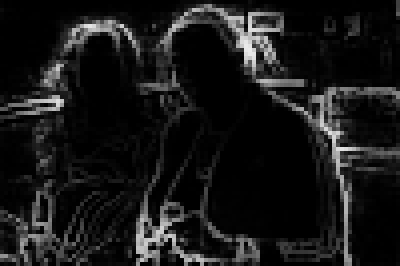}
\includegraphics[width=.09\textwidth]{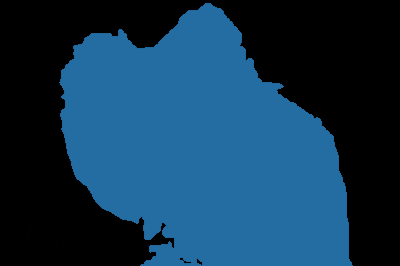}
\includegraphics[width=.09\textwidth]{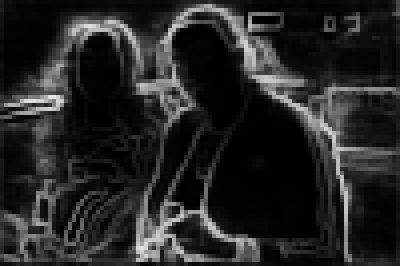}
\includegraphics[width=.09\textwidth]{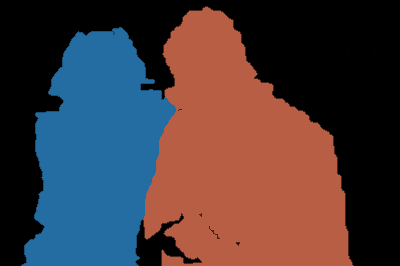}
\\
\centering
\begin{minipage}[t]{0.09\textwidth}
\centering
\scriptsize{\textbf{input\\image}}
\end{minipage}
\begin{minipage}[t]{0.09\textwidth}
\centering
\scriptsize{\textbf{boundary\\(w/o BE)}}
\end{minipage}
\begin{minipage}[t]{0.09\textwidth}
\centering
\scriptsize{\textbf{instance mask\\(w/o BE)}}
\end{minipage}
\begin{minipage}[t]{0.09\textwidth}
\centering
\scriptsize{\textbf{boundary\\(w/ BE)}}
\end{minipage}
\begin{minipage}[t]{0.09\textwidth}
\centering
\scriptsize{\textbf{instance mask\\(w/ BE)}}
\end{minipage}
\caption{Visualization of the boundary detection branch with and without the BE module, \kk{which shows the effectiveness of our proposed BE module in mining continuous boundary information for separating salient objects of same classes.} 
}
\label{fig:wbe}
\vspace{-3mm}
\end{figure}

Figure~\ref{fig:wbe} visualizes two examples of boundary detection and the corresponding salient instance detection with and without the BE module. We can see that our BE module helps detect the boundaries between objects, which is crucial to salient instance segmentation.

\subsubsection{Network Structure}
The top part of Figure~\ref{fig:pipeline} shows the architecture of the boundary detection branch. Given an input image $\mathcal{I}$, the backbone network produces multi-scale features ($f_{1}$ to $f_{5}$), each of which is enhanced by {a CA module} (to be discussed in Section~\ref{sec:da}) before they are concatenated and computed to predict the boundary map. {We also feed the input image into the BE module to obtain enhanced edge features $f_{b}$.} The output boundary map $\mathcal{B}$ is then computed as:
$\mathcal{B} = \sigma(Conv(Concat({f_{1}}^{*},...,{f_{5}}^{*},f_{b1},f_{b2})))$, where $\sigma$ is the sigmoid activation function.

\subsection{Cross-layer Attention (CA) Module}\label{sec:da}
\begin{figure}[tb]
\includegraphics[width=0.9\linewidth]{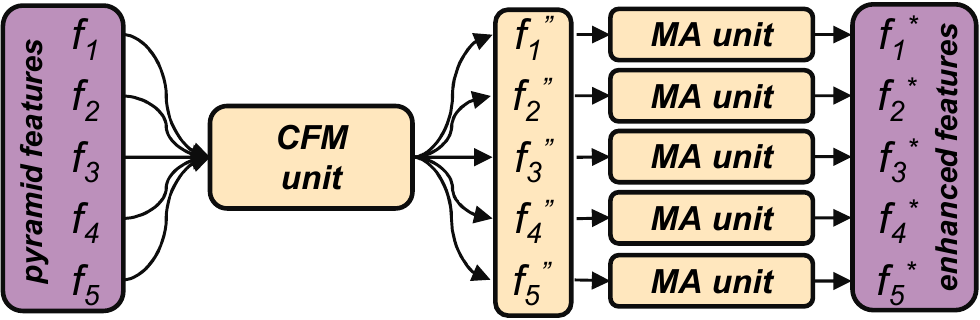}
\vspace{-3mm}
\centering
\caption{\tx{Cross-layer Attention (CA) module.}}
\label{fig:CA}
\end{figure}

Detecting instance centroids and boundaries are two highly coupled sub-tasks, \ie, they can influence each other and further affect the SID performance. To \kk{effectively} learn these two sub-tasks, we propose the \tx{Cross-layer Attention (CA)} module \kk{for refining backbone features before they are used for these two sub-tasks.} Its design is based on \kk{two} observations.
%
\kk{First, low-level features contain high-resolution but noisy information for delineating salient instance boundaries, while high-level features have low-resolution but robust information for salient instance localization.}
\kk{Second, since salient instances may have various shapes and they may correspond to different class labels, we need to model both long-range spatial and cross-channel contextual information.}
%
%
\kk{Unlike existing dual attention mechanisms~\cite{woo2018cbam,fu2019dual} that only enhance the feature representation capacity of one fixed layer, our CA module first incorporates a Cross-layer Feature Mixing (CFM) unit to enhance the communication across different levels of backbone features and then uses multiple Mutual Attention (MA) units to learn hierarchical channel-wise and spatial-wise attentive features for each sub-task. 
\txx{
The CFM unit shares its parameters to allow information exchanges across the boundary and centroid branches.
}
%
Figure~\ref{fig:CA} shows the module structure.
}


\subsubsection{Structure of CFM Unit}
\begin{figure}[h]
\includegraphics[width=0.9\linewidth]{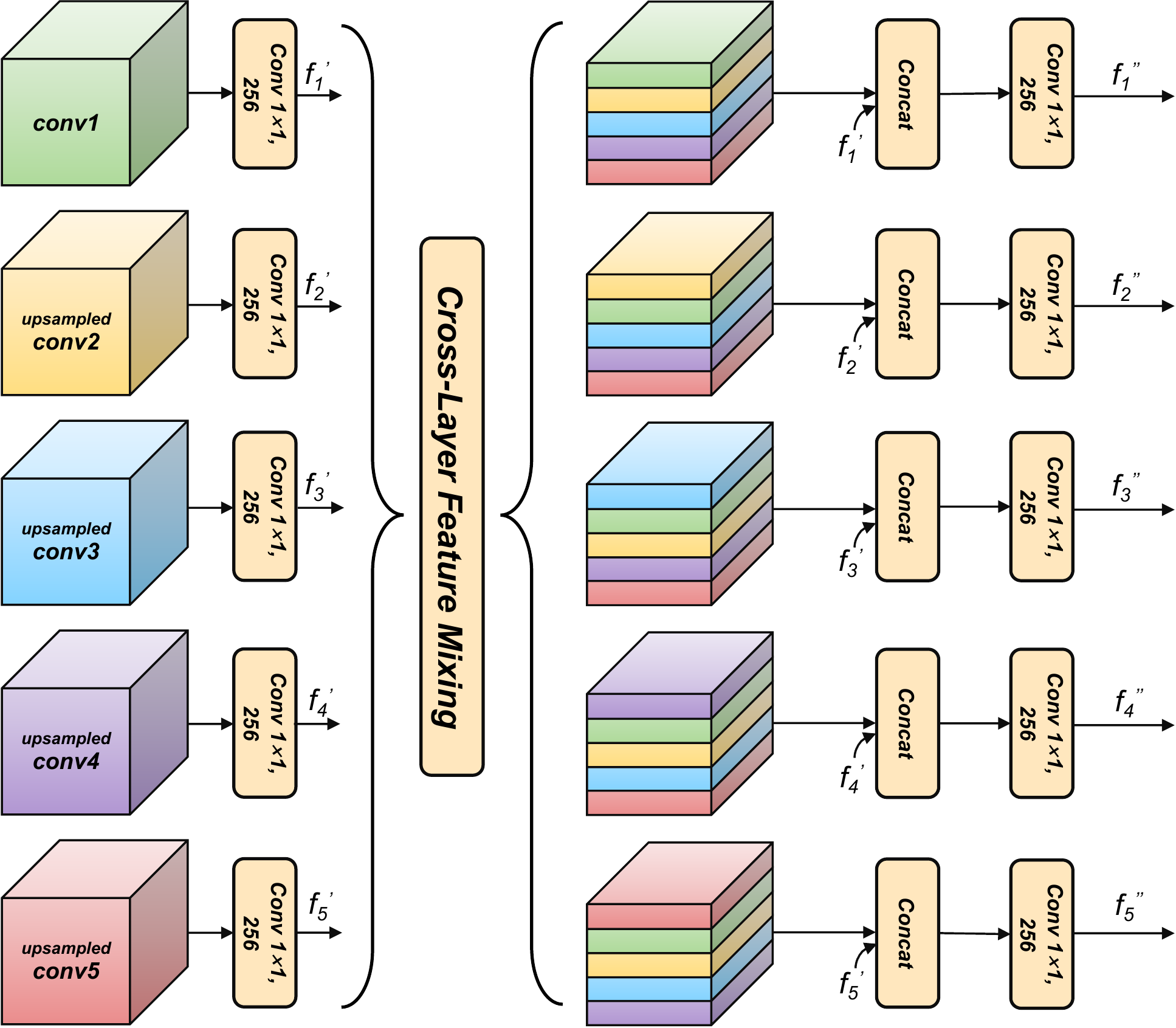}
\vspace{-1mm}
\centering
\caption{\txx{Cross-layer Feature Mixing (CFM) unit.}}
\label{fig:CL}
\vspace{-3mm}
\end{figure}

Figure~\ref{fig:CL} shows the structure of the CFM unit. $f_1$ to $f_5$ are the features from the pyramid layers of the ResNet backbone. We first upsample \kkk{$f_2$}, $f_3$, $f_4$, and $f_5$ such that the given feature maps have the same resolution, and apply $1\times1$ convolution on the five feature maps such that they have the same channel depth (256). We get features ${f_{1}}^{'}$ to ${f_{5}}^{'}$ that have the same shape for the following operation. We then apply CFM on the pyramid features to generate cross-layer features. CFM is implemented via a concatenation-split-concatenation operation on the feature channels. We concatenate features ${f_{1}}^{'}$ to ${f_{5}}^{'}$ as $f_c$, with 1280 (256$\times$5) channels. The split-concatenation operation could be considered as a reshape-transpose-reshape process. We reshape channel dimension of $f_c$ to 2 dimensions (\ie,~[5, 256]), transpose it to [256, 5], and then flatten it to 1280.
Finally, we concatenate the features before/after CFM, and feed these concatenated features to 1$\times$1 convolutional filters to generate the final enhanced features (${f_1}^{''}$ to ${f_5}^{''}$).

\subsubsection{Structure of MA Unit}
\begin{figure}[h]
\begin{center}
\includegraphics[width=0.95\linewidth]{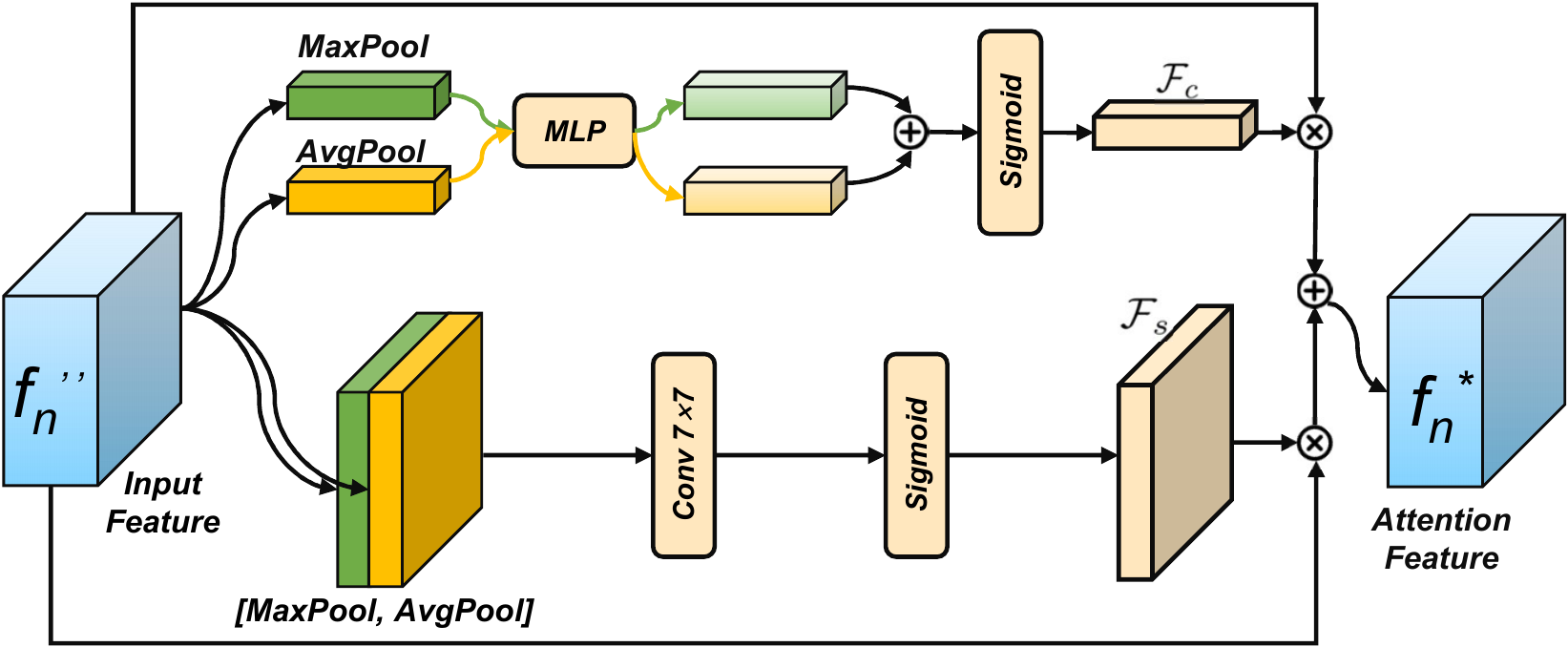}
\end{center}
\vspace{-3mm}
\centering
\caption{\tx{Mutual Attention (MA) unit.}}
\label{fig:MA}
\vspace{-3mm}
\end{figure}

Figure~\ref{fig:MA} shows the structure of our MA unit. The top and bottom branches are channel-wise and spatial-wise attention blocks, respectively. Specifically, given the input features ${{f}_{n}}^{''}$, we compute the channel-wise attention features $\mathcal{F}_{c}$ as: 
\begin{equation}
\footnotesize{\mathcal{F}_{c} =  \sigma(MLP(AvgPool_{c}({{f}_{n}}^{''})) +MLP(MaxPool_{c}({{f}_{n}}^{''}))),}
\label{equa:1}
\end{equation}
where $MaxPool_{c}$ and $AvgPool_{c}$ denote two channel-wise pooling operations, and MLP is the multi-layer perception with one hidden layer to generate the attention features. We also compute the spatial-wise attention features $\mathcal{F}_{s}$ as:
\begin{equation}
\footnotesize{\mathcal{F}_{s} = \sigma(Conv_{7\times7}([AvgPool_{s}({{f}_{n}}^{''});MaxPool_{s}({{f}_{n}}^{''})])),}
\label{equa:2}
\end{equation}
where $Conv_{7\times7}$ is a convolutional layer with kernel size 7. The final attention features ${{f}_{n}}^{*}$ are then computed as:
\begin{equation}
{{f}_{n}}^{*} = {{f}_{n}}^{''}\times\mathcal{F}_{c} + {{f}_{n}}^{''}\times\mathcal{F}_{s},
\end{equation}
where $\times$ denotes the dot product operation, and $+$ is the element-wise summation operation.

\vspace{3mm}

\begin{figure}[!b]
\centering
\includegraphics[width=.1\textwidth]{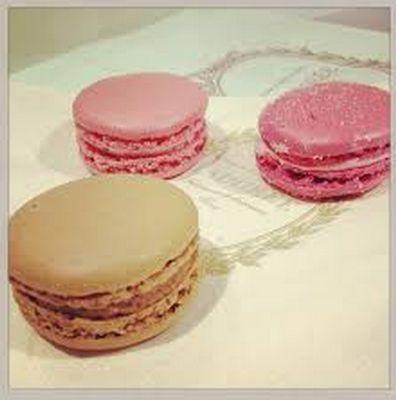}
\includegraphics[width=.1\textwidth]{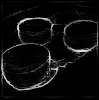}
\includegraphics[width=.1\textwidth]{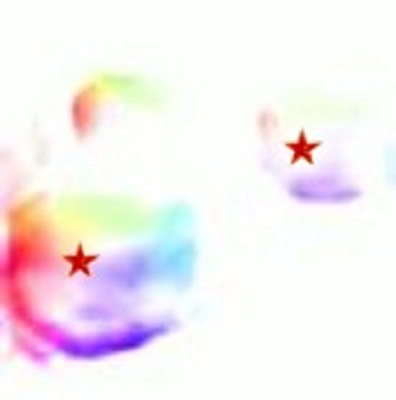}
\includegraphics[width=.1\textwidth]{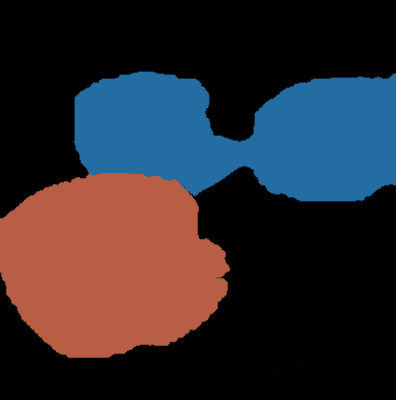}
\\
\centering
\begin{minipage}[t]{0.1\textwidth}
\centering
\scriptsize{\textbf{input\\ image}}
\end{minipage}
\begin{minipage}[t]{0.1\textwidth}
\centering
\scriptsize{\textbf{boundaries \\(w/o CA)}}
\end{minipage}
\begin{minipage}[t]{0.1\textwidth}
\centering
\scriptsize{\textbf{centroids \\ (w/o CA)}}
\end{minipage}
\begin{minipage}[t]{0.1\textwidth}
\centering
\scriptsize{\textbf{instance mask \\(w/o CA)}}
\end{minipage}
\\

\vspace{0.1cm}
\centering
\includegraphics[width=.1\textwidth]{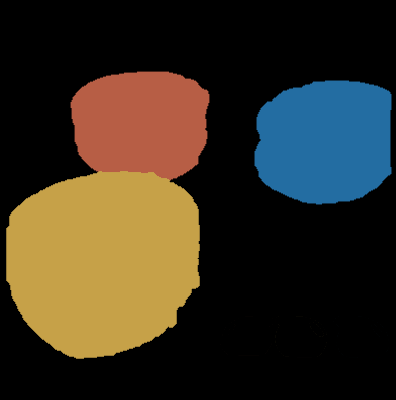}
\includegraphics[width=.1\textwidth]{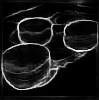}
\includegraphics[width=.1\textwidth]{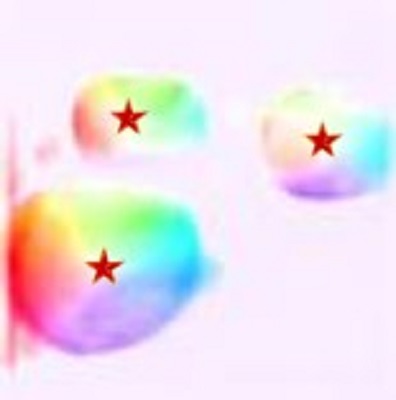}
\includegraphics[width=.1\textwidth]{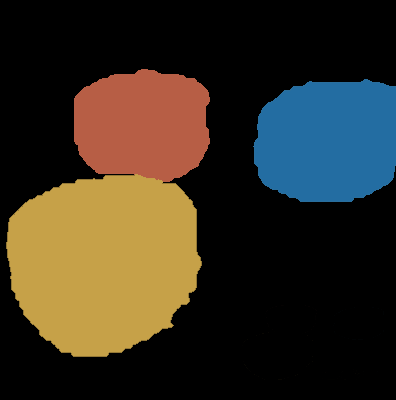}
\vspace{0.1cm}\\
\centering
\begin{minipage}[t]{0.1\textwidth}
\centering
\scriptsize{\textbf{ground\\ truth}}
\end{minipage}
\begin{minipage}[t]{0.1\textwidth}
\centering
\scriptsize{\textbf{boundaries \\(w/ CA)}}
\end{minipage}
\begin{minipage}[t]{0.1\textwidth}
\centering
\scriptsize{\textbf{centroids \\ (w/ CA)}}
\end{minipage}
\begin{minipage}[t]{0.1\textwidth}
\centering
\scriptsize{\textbf{instance mask \\(w/ CA)}}
\end{minipage}
\caption{\kkk{Illustration on how the} CA module benefits both boundary and centroid tasks. \tx{We can see that the CA module influence both the continuity of the detected boundaries and the accuracy of detected centroids.}}\label{fig:da}
\vspace{-4.5mm}
\end{figure}

Figure~\ref{fig:da} shows the effectiveness of the proposed CA module in enhancing the boundary and centroid detection performances.

\begin{figure*}[!t]
\centering
\includegraphics[width=0.99\textwidth]{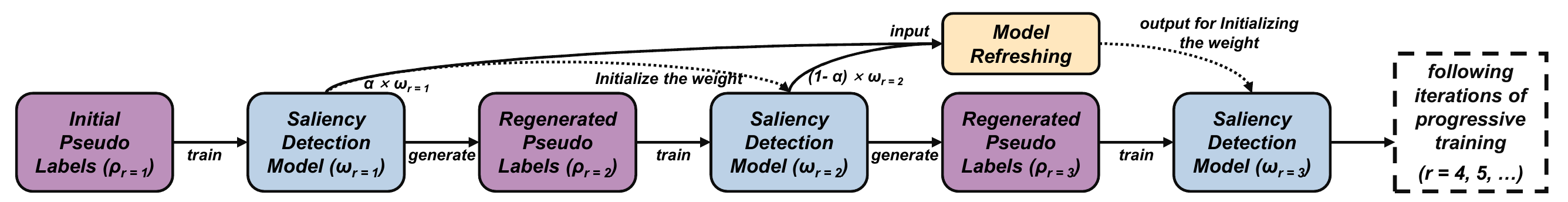}
\vspace{-3mm}
\caption{\tx{Progressive Training Scheme for our saliency detection branch. Each training iteration in the dashed box contains two steps, model refreshing and pseudo label regeneration, leading the model to correct the noise in a self-supervised manner.}}\label{fig:pts}
\vspace{-3mm}
\end{figure*}

\subsection{Progressive Training Scheme (PTS)}
%
\kk{Weak annotated labels would inevitably introduce noise into the learning process. To reduce the noise, previous works propose to use temporal ensemble learning~\cite{laine2016temporal,tarvainen2017mean} in a semi-supervised setting, where latent knowledge learned from labeled data can be applied to noisy unlabeled data. We extend this idea in our weakly-supervised setting. Since we do not have fully-annotated data, we explore this ensemble learning strategy in a progressive self-supervised manner, \ie, by reciprocally training the salient object detection branch using newly predicted salient instance maps. It mainly contains two iterative steps: pseudo label generation, and model refreshing. Figure~\ref{fig:pts} shows the overview of the proposed Progressive Training Scheme, and Algorithm~\ref{alg:1} summarizes the main steps.}
%
%

\subsubsection{Pseudo Label Generation}
\kk{Since we do not have any fully-annotated labels to learn a noise-free representation, we propose to use our WSID-Net as a pseudo-label generator and refine its salient object detection branch in a self-supervised manner. This is because the output of our WSID-Net has more accurate boundaries with the help of other two branches, compared with its salient object detection branch. On the other hand, the re-trained saliency detection branch can further boost the performance of our WSID-Net due to the improved salient object localization. To this end, we first forward the WSID-Net to generate salient instance masks, and use them as the initial pseudo training labels to update the network parameters of the salient object detection branch. In the next training iteration, we update the pseudo training labels using the re-trained WSID-Net.}


\subsubsection{Model Refreshing}
\kk{Before we update the pseudo training labels in the next iteration, we need to refresh our salient object detection branch. Note that we do not have any clean data (fully-annotated labels in our case) that can be used for learning noise-free features. It is possible that our model may overfit the noise in the pseudo labels and converge to a local minimum.}
%
\kk{To avoid these problems, we adopt the Exponential Moving Average method to update the model parameters with a weighted sum of model parameters in the current and former iterations. We define ${\omega}_{r}$ as the model weight in iteration $r$, and the model refreshing is formulated as}:
\begin{equation}
{\omega}_{r+1} = \alpha{\omega}_{r-1} + (1-\alpha){\omega}_{r},
\label{equa:1}
\end{equation}
where $\alpha$ is the smoothing hyper-parameter \kk{that balances the contributions of model parameters from different iterations}.

\begin{algorithm}[t]
\floatname{algorithm}{Algorithm}
\caption{: Progressive Training Scheme (PTS).}~\label{alg:1}
\begin{algorithmic}[1]
\Require Initial pseudo label ${\rho}_{1}$, initial model weight ${\omega}_{1}$, training iterations $R$, and training epochs $E$ per iteration
\Ensure Final trained model weight ${\omega}_{f}$
\For{$r=1$ to $R$}
\For{$e=1$ to $E$}
\State update model weight ${\omega}_{r}$ via backpropagating gradients and learning from pseudo labels ${\rho}_{r}$
\EndFor
\If{\txx{$r > 2$}}
\State  update model weight ${\omega}_{r}$ using Eq.~\ref{equa:1}
\EndIf
\State generate pseudo label ${\rho}_{r+1}$ using WSID-Net, where the saliency branch uses model weight ${\omega}_{r}$, and further uses CRF to refine the boundaries of the pseudo labels
\EndFor
\State ${\omega}_{f} = {\omega}_{R}$
\end{algorithmic}
\end{algorithm}

\begin{figure}[!tp]
\centering
\includegraphics[width=.1\textwidth]{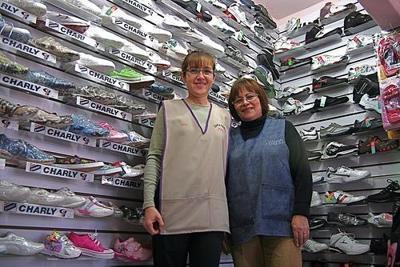}
\includegraphics[width=.1\textwidth]{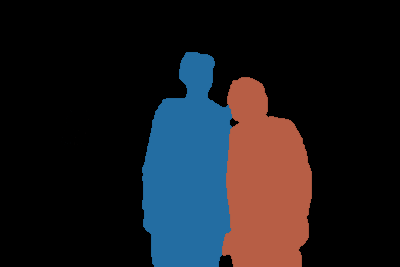}
\includegraphics[width=.1\textwidth]{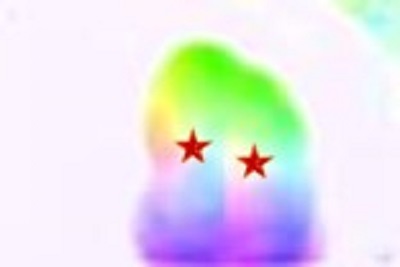}
\includegraphics[width=.1\textwidth]{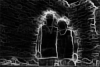}
\centering
\begin{minipage}[t]{0.1\textwidth}
\centering
\scriptsize{{(a) Input}}
\end{minipage}
\begin{minipage}[t]{0.1\textwidth}
\centering
\scriptsize{{(b) Ground truth}}
\end{minipage}
\begin{minipage}[t]{0.1\textwidth}
\centering
\scriptsize{{(c) Centroids}}
\end{minipage}
\begin{minipage}[t]{0.1\textwidth}
\centering
\scriptsize{{(d) Boundaries}}
\end{minipage}
\centering
\includegraphics[width=.1\textwidth]{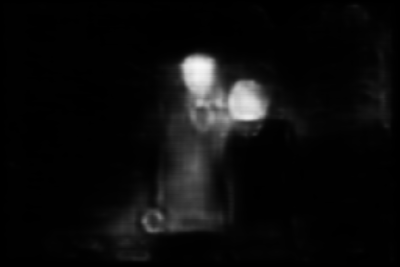}
\includegraphics[width=.1\textwidth]{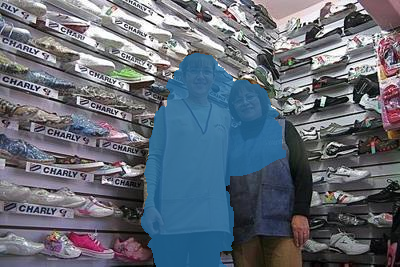}
\includegraphics[width=.1\textwidth]{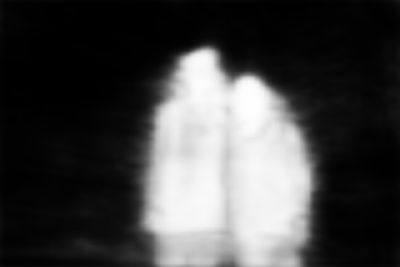}
\includegraphics[width=.1\textwidth]{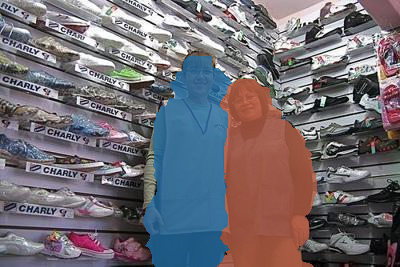}
\begin{minipage}[t]{0.1\textwidth}
\centering
\scriptsize{{(e) Saliency map (w/o PTS)}}
\end{minipage}
\begin{minipage}[t]{0.1\textwidth}
\centering
\scriptsize{{(f) Instance mask \\ (w/o PTS) }}
\end{minipage}
\begin{minipage}[t]{0.1\textwidth}
\centering
\scriptsize{{(g) Saliency map \\ (w/ PTS)}}
\end{minipage}
\begin{minipage}[t]{0.1\textwidth}
\centering
\scriptsize{{(h) Instance mask \\ (W/ PTS)}}
\end{minipage}
\caption{\kk{Illustration of how the proposed PTS benefits our task progressively in a self-supervised manner. While WSID-Net initially fails due to inaccurate detection (e), the output of WSID-Net (f) is still better than (e) with the help of the centroid (c) and boundary (d) detections. By using (f) to further refine the saliency detection branch (g), our WSID-Net could also enjoy the performance boost (h).}}
\label{fig:PTS_example}
\end{figure}

%
\kk{Figure~\ref{fig:PTS_example} shows one example of how the proposed PTS benefits our task in a self-supervised manner: given a challenging input image of two persons leaning on each other (a), our saliency detection branch fails to detect the majority of the salient regions due to the diverse foreground appearance and cluttered background (e), which further causes the failure of our WSID-Net (f). However, since (f) is still better than (e) due to the help of the centroid and boundary detections ((c) and (d)), after we have applied the proposed PTS training strategy, we can see that our saliency detection branch enjoys a better performance in detecting the saliency region as a whole (g), leading to an accurate salient instance detection result (h).}
%
%


\section{Experiments}

\subsection{Implementation \kk{Details}}
\tx{We implement WSID-Net on the Pytorch framework~\cite{paszke2019pytorch}.
Both training and testing are performed on a PC with an i7 4GHz CPU and a GTX 1080Ti GPU.
CRF is used to generate or refine pseudo labels. The hyper parameters of CRF are set as $w_{1}=4.0$, $w_{2}=3.0$, $\sigma_{\alpha}=49.0$, $\sigma_{\beta}=5.0$, and $\sigma_{\gamma}=3.0$.
We choose ResNet50 as the backbone for all three branches in WSID-Net. The backbone are initialized as in \cite{simonyan2014very}.
Input images are resized to 512$\times$512 resolution.
To minimize the loss function, we use the SGD optimizer with batch size 6 and initial learning rate 0.01. The learning rate decreases following poly policy ($lr_{itr} = lr_{init}(1-\frac{itr}{max_itr})^{\gamma}$). We train our WSID-Net for 5 epoches.}
\kk{The proposed PTS training is further applied to refine the saliency detection branch for another 6 iterations, of which each iteration contains 8 epoches ($R$ and $E$ in Algorithm~\ref{alg:1}). $\alpha$ in Eq.~\ref{equa:1} is set to $\frac{r}{r+1}$, where $r$ is the current iteration index. The learning rate begins with 0.0001, and the decay follows the aforementioned poly policy.
}

\begin{table*}[!t]
\vspace{6mm}
\caption{Quantitative evaluation of our method against six baseline methods and state-of-the-art fully-supervised SID methods. For the compared methods, we show their original tasks, supervision types, training labels and auxiliary pre-trained models in the 2$nd$ to 5$th$ columns. SID, SOD, OD, and SIS represent salient instance detection, salient object detection, object detection and semantic instance segmentation, respectively. FS and WS denote Fully-Supervised and Weakly-Supervised. Best performances among the weakly-supervised methods are marked in \textbf{\textcolor{red}{red}}.
}%
\vspace{-3mm}
\begin{center}
 \begin{tabular}{|c c c c c c c|}
 \hline
 \textbf{{Methods}} & \textbf{\begin{tabular}[c]{@{}c@{}}{Original} \\ {task}\end{tabular}} & \textbf{\begin{tabular}[c]{@{}c@{}}{Supervision} \\ {types}\end{tabular}} & \textbf{\begin{tabular}[c]{@{}c@{}}{Training labels}\end{tabular}} & \textbf{\begin{tabular}[c]{@{}c@{}}{Auxiliary} \\{models}\end{tabular}} & \textbf{\begin{tabular}[c]{@{}c@{}}{mAP} \\ {@0.5$\uparrow$}\end{tabular}} & \textbf{\begin{tabular}[c]{@{}c@{}}{mAP} \\ {@0.7$\uparrow$}\end{tabular}} \\ 
 \hline\hline
 {MSRNet \cite{li2017instance}} & {SID} & {FS} & \begin{tabular}[c]{@{}c@{}}{object-level and} \\{instance-level pixel masks}\end{tabular} & {MAP \cite{zhang2016unconstrained}, MCG \cite{APBMM2014}} & {{65.3\%}} & {{52.3\%}} \\
 \hline
 {MAP \cite{zhang2016unconstrained}} & {SID} & \scriptsize{FS}  & {instance-level bounding boxes} & {N/A} & {56.6\%} & {24.8\%} \\
 \hline
 {S4Net \cite{fan2019s4net}} & {SID} & {FS} & {instance-level pixel masks} & {N/A} & {{82.2\%}} & {{59.6\%}} \\
 \hline
 \hline
 {C2SNet \cite{li2018contour}} & {SOD} & {WS} & {unlabeled images} & \begin{tabular}[c]{@{}c@{}}{CEDN \cite{Yang2016object},} \\ {MAP \cite{zhang2016unconstrained},} {MCG \cite{APBMM2014}}\end{tabular} & {41.1\%} & {25.4\%} \\
 \hline
 {NLDF \cite{luo2017non}} & {SOD} & {WS} & \begin{tabular}[c]{@{}c@{}}{object-level pixel masks}\end{tabular} & {MAP \cite{zhang2016unconstrained}, MCG \cite{APBMM2014}} & {45.5\%} & {24.5\%} \\
 \hline
 {DeepMask \cite{pinheiro2015learning}} & {OD} & {WS} & {instance-level bounding boxes} & {N/A} & {37.1\%} & {20.5\%} \\
 \hline
 {PRM+D \cite{cholakkal2019object}} & {SIS} & {WS} & {class, subitizing labels} & {MCG \cite{APBMM2014}} & {49.6\%} & {31.2\%} \\
 \hline
 {IRN \cite{jiwoon2019weakly}} & {SIS} & {WS} & {class labels} & {N/A} & {57.1\%} & {37.4\%} \\
 \hline
 {Ours} & {SID} &  {WS}  & {class, subitizing labels} & {N/A} &   \textbf{{\textcolor{red}{68.3\%}}}  &  \textbf{{\textcolor{red}{51.7\%}}} \\
 \hline
\end{tabular}
\end{center}
\label{tab:SID}
\end{table*}

\subsection{Training and Evaluation Details}
\paragraph{Datasets.}
\kk{Our WSID-Net is trained on two kinds of image-level labels, class and subitizing. We use the class labels from the PASCAL VOC 2012~\cite{Everingham2010-pascal-voc-IJCV} dataset (which is originally proposed for semantic and instance segmentation) to train our network. For subitizing labels, we count the numbers of salient instances from the ILSO~\cite{li2017instance} dataset, and train our network on this training set. For the proposed progressive training scheme, we augment the training data by using the unlabeled training data from C2SNet~\cite{li2018contour}. For testing, following the existing fully-supervised SID method~\cite{fan2019s4net}, we perform \tx{SID} evaluations on the test set of ILSO~\cite{li2017instance}.}



\paragraph{Training and Inference.}
\txx{We train the boundary and centroid branches together with different losses, and train the saliency branch independently.}
%
We train the centroid detection branch using the proposed centroid-based subitizing loss together with the centroid loss introduced in~\cite{jiwoon2019weakly,Neven2019InstanceSB}.
We train the boundary detection branch using the boundary loss introduced in~\cite{Ahn_2018_CVPR,jiwoon2019weakly}.
%
%
%
To train the saliency detection branch, we follow existing weakly-supervised SOD methods to use pseudo masks derived from class labels. Specifically, we first compute class activation maps via~\cite{zhou2016learning}. We then feed these maps together with the input image to a Conditional Random Field (CRF)~\cite{krahenbuhl2011efficient} to generate pseudo object maps, and use these pixel-level pseudo labels to train the saliency detection branch. \tx{We further utilize the proposed PTS with model refreshing and self-generated pseudo labels to retrain the saliency detection branch. }

During inference, given an input image, WSID-Net first computes the centroids, boundaries, and saliency maps.
\txx{We first obtain the initial saliency instance map $\mathcal{SI}^{*}$ via the saliency map and centroid map, as discussed in Section~\ref{lsu_loss}. We then use the boundary map to refine the initial saliency instance map with the random walk algorithm.
The transition probability matrix $\mathcal{M}$ is defined as:
\begin{equation}
    \mathcal{M} = \mathcal{D}^{-1}\mathcal{H}^{\chi},
\end{equation}
where $\mathcal{H}$ is the affinity matrix of the learned boundary map $\mathcal{B}$, and $\mathcal{D}$ is a diagonal matrix relating to $\mathcal{H}$. The element in $\mathcal{H}$ is defined as: $h_{k} = 1 - \max \limits_{k \in \Pi_{ij}}\mathcal{B}(x_{k})$, where $\Pi_{ij}$ is a set of pixels on the line between boundary pixels $x_{i}$ and $x_{j}$. In addition, $\mathcal{H}^{\chi}$ is the self production of $\mathcal{H}$ with power $\chi$ for affinity distillation, and $\mathcal{D}$'s diagonal element $\mathcal{D}_{ii}$ equals to $\sum{h}^{\chi}_{ij}$ for summarizing values of $\mathcal{H}^{\chi}$ by row. 
%
The random walk algorithm for instance-wise saliency value propagation is conducted as:
\begin{equation}
    vec(\overline{\mathcal{SI}}^{*}_{n}) = \mathcal{M}^{i}vec(\mathcal{SI}^{*}_{n}(1-\mathcal{B})),
\end{equation}
where $vec()$ refers to the vectorization of the matrix, and $\overline{\mathcal{SI}}^{*}_{n}$ is our final saliency instance map. 
}


\paragraph{Evaluation Metrics.}
We use the mean Average Precision (mAP) metric~\cite{hariharan2014simultaneous} to evaluate the SID performance. The IoU is set to 0.5 and 0.7 for this metric.

\subsection{\kk{Comparing to the State-of-the-art Methods}}\label{sec42}
As we are the first to propose a weakly-supervised SID method, \kk{we compare our method to 2 existing fully-supervised state-of-the-art SID methods: S4Net~\cite{fan2019s4net} and MSRNet~\cite{li2017instance}.
We also prepare the following baselines from related tasks for evaluation. We choose 6 state-of-the-art weakly-supervised methods, with two from the SOD task C2SNet \cite{li2018contour} and NLDF~\cite{luo2017non}; one from the SID task MAP~\cite{zhang2016unconstrained}; one from the object detection (OD) task, DeepMask \cite{pinheiro2015learning}; and two from the Semantic Instance Segmentation task, PRM+D~\cite{cholakkal2019object} and IRN \cite{jiwoon2019weakly}. We adapt them by adding different post-processing strategies to these methods for deriving instance-level saliency maps from their original outputs, or modifying their networks and retrain them using our training data. Details are summarized as follows:}
\begin{itemize}
\item \txx{We choose ``MCG~\cite{APBMM2014} + MAP~\cite{zhang2016unconstrained}" as the post-processing strategy for the weakly-supervised SOD methods (\ie, C2SNet~\cite{li2018contour} and NLDF~\cite{luo2017non}), inspired by the fully-supervised SID method MSRNet \cite{li2017instance}. It has been shown in \cite{li2017instance} that MCG~\cite{APBMM2014} can be used to produce segments given the contour maps as input and assign these segments with confidence scores. Segments with low confidences can then be filtered out by MAP~\cite{zhang2016unconstrained}. Considering that both C2SNet~\cite{li2018contour} and NLDF~\cite{luo2017non} learn to produce contour maps, we find this post-process strategy suitable for weakly-supervised SOD methods with contour predictions.
}

\item \txx{We select CRF~\cite{krahenbuhl2011efficient} as the post-processing strategy for fully-supervised bounding-box-based SID method MAP~\cite{zhang2016unconstrained}, due to the fact that CRF is a popular graphical model used as post-processing for boosting segmentation performance. Considering that MAP~\cite{zhang2016unconstrained} can generate instance-level bounding boxes, CRF can be used to obtain instance maps by refining the boundaries, which gives a performance boost of $3\%$.
}

\item \txx{We choose a weakly-supervised SOD method as post-processing for filtering out non-salient segments produced by DeepMask~\cite{pinheiro2015learning}, as DeepMask~\cite{pinheiro2015learning} is a class-agnostic object detection method that is not aware of saliency information. We choose WSS~\cite{wang2017learning} as the weakly-supervised SOD method for a fair comparison, as it performs closely to our Saliency Detection Branch in our preliminary experiment. This strategy improves the performance by $5\%$.
}

\item \txx{IRN~\cite{jiwoon2019weakly} produces class-specific instance segmentation maps, which do not have saliency information. To adapt its results from class-specific to class-agnostic, we remove the CAM in their method and directly use their centroid and boundary maps to obtain instance maps. We then utilize WSS~\cite{wang2017learning} to select salient instances.
}

\item \txx{PRM+D~\cite{cholakkal2019object} is trained with class and per-class subitizing labels to predict semantic instance maps. However, this method can only response to the instances with pre-defined class labels. To adapt it to class-agnostic, we merge its per-class outputs (originally 20 output maps for 20 classes) into one class-agnostic map by adding an additional convolutional layer, and then retrain it using our training data.
}
\end{itemize}

\paragraph{Quantitative Comparisons.} \kk{We quantitatively evaluate our method in Table~\ref{tab:SID}$^{\dagger}$. It is worth noting that} our method achieves a significantly better performance of about \tx{20\%} over the second-place weakly-supervised baseline, \kk{on the mAP@0.7 metric} (which is very challenging as it requires the IoU value to be over 70\%). These results show that our method achieves the best performance using just two image-level labels.

{\let\thefootnote\relax\footnote{{$^{\dagger}$ As of today, the codes for MSRNet~\cite{li2017instance} are still not available. Following~\cite{fan2019s4net}, we directly copy the numbers reported in~\cite{li2017instance} to our submission for a quantitative comparison.}}}

\begin{figure*}[!pt]
\centering
\begin{minipage}[t]{0.040\textwidth}
\centering
\tiny{\textbf{}}
\end{minipage}
\centering
\begin{minipage}[t]{0.090\textwidth}
\centering
\scriptsize{\textbf{Image}}
\end{minipage}
\begin{minipage}[t]{0.090\textwidth}
\centering
\scriptsize{\textbf{PRM+D \cite{cholakkal2019object}}}
\end{minipage}
\begin{minipage}[t]{0.087\textwidth}
\centering
\scriptsize{\textbf{DeepMask\cite{pinheiro2015learning}}}
\end{minipage}
\begin{minipage}[t]{0.087\textwidth}
\centering
\scriptsize{\textbf{C2SNet \cite{li2018contour}}}
\end{minipage}
\begin{minipage}[t]{0.087\textwidth}
\centering
\scriptsize{\textbf{IRN \cite{jiwoon2019weakly}}}
\end{minipage}
\begin{minipage}[t]{0.087\textwidth}
\centering
\scriptsize{\textbf{NLDF \cite{luo2017non}}}
\end{minipage}
\begin{minipage}[t]{0.087\textwidth}
\centering
\scriptsize{\textbf{MAP \cite{zhang2016unconstrained}}}
\end{minipage}
\begin{minipage}[t]{0.087\textwidth}
\centering
\scriptsize{\textbf{S4Net \cite{fan2019s4net}}}
\end{minipage}
\begin{minipage}[t]{0.087\textwidth}
\centering
\scriptsize{\textbf{WSID-Net (Ours)}}
\end{minipage}
\centering
\begin{minipage}[t]{0.087\textwidth}
\centering
\scriptsize{\textbf{GT}}
\end{minipage}
\\
\centering
\includegraphics[width=.95\textwidth]{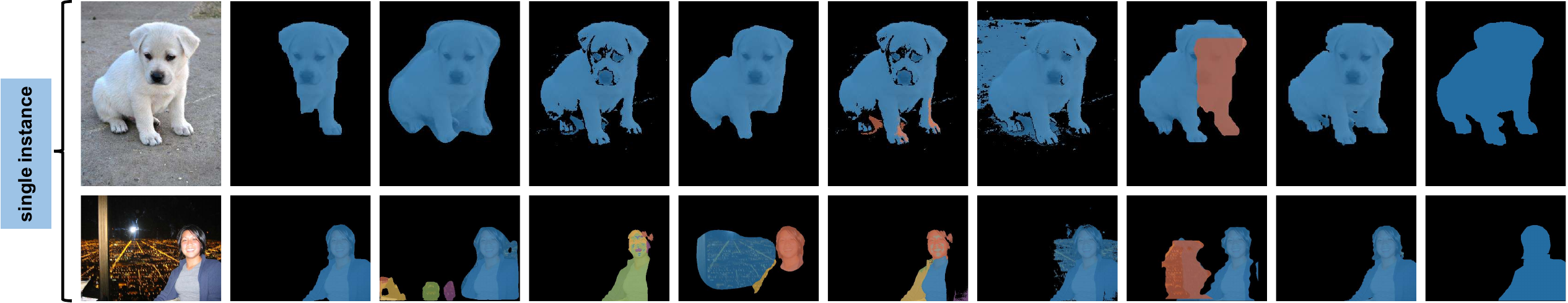}
\\
\centering
\includegraphics[width=.95\textwidth]{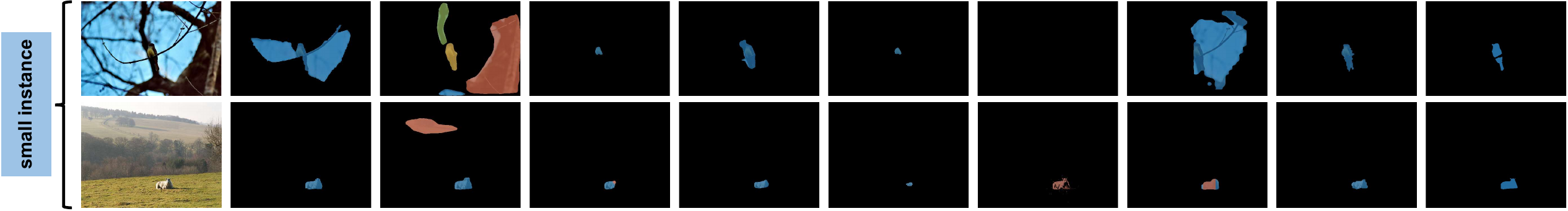}
\\
\centering
\includegraphics[width=.95\textwidth]{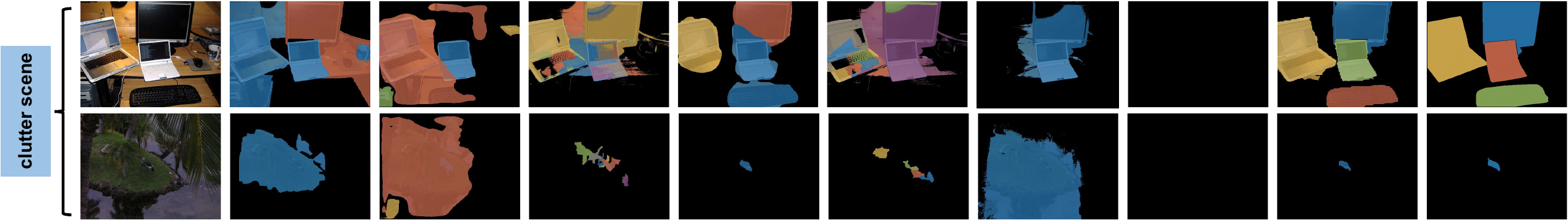}
\\
\centering
\includegraphics[width=.95\textwidth]{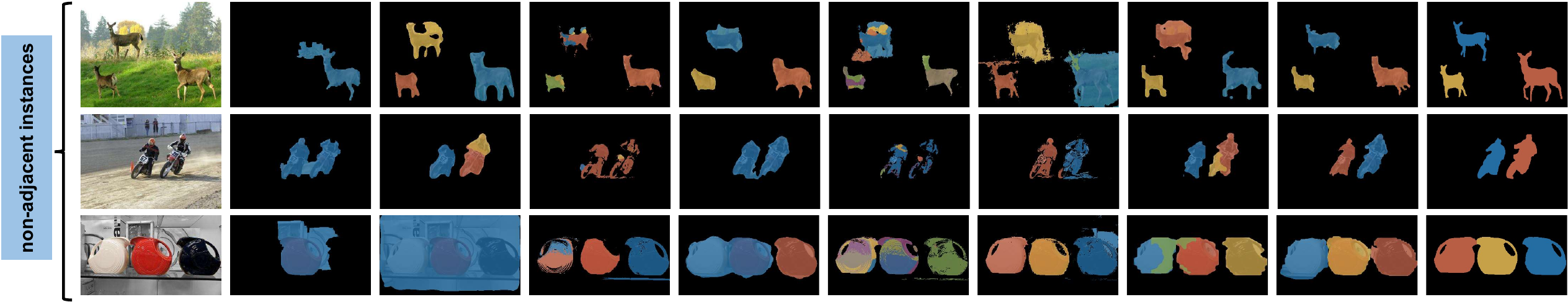}
\\
\centering
\includegraphics[width=.95\textwidth]{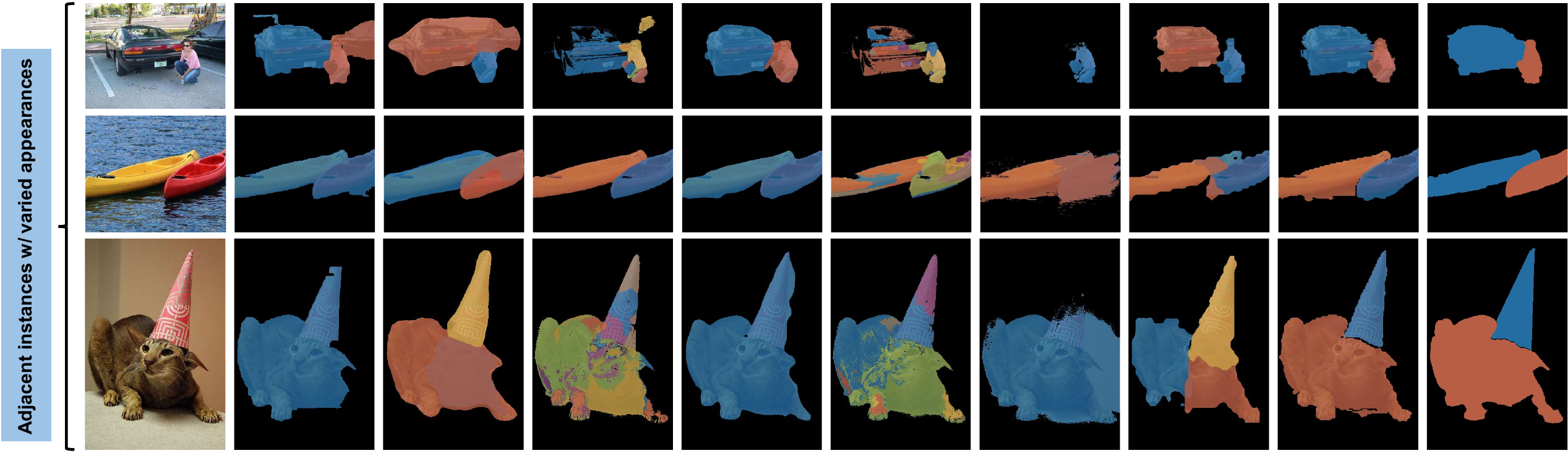}
\\
\centering
\includegraphics[width=.95\textwidth]{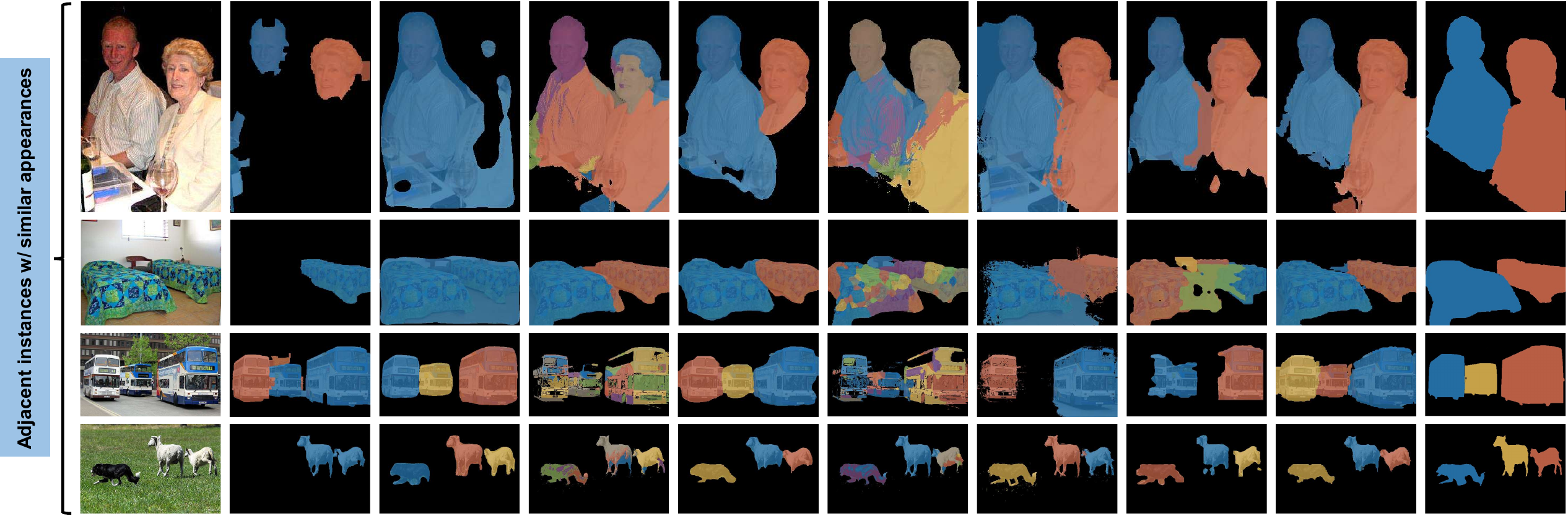}
\\
\caption{Qualitative results of our method, compared with existing fully-supervised methods (S4Net\cite{fan2019s4net} and MAP~\cite{zhang2016unconstrained}) and modified baselines (PRM+D~\cite{cholakkal2019object}, DeepMask~\cite{pinheiro2015learning}, C2SNet~\cite{li2018contour}, NLDF~\cite{luo2017non}, and IRN~\cite{jiwoon2019weakly}). Refer to Section~\ref{sec42} and Table~\ref{tab:SID} on how we modify and train these baselines, in order to carry out a fair comparison. }
\label{fig:instance}
\end{figure*}

\paragraph{Qualitative Comparisons.} We further qualitatively evaluate our method with fully-supervised methods and baselines in Figure~\ref{fig:instance}. \tx{The visual results verify that our method is able to delineate the instance boundaries clearly, and output accurate numbers of segmented salient instances directly for different scenes, \ie~scenes with single instances, small instances, (non-)adjacent instances, similar/varied instances, and cluttered contents. In contrast, the compared methods exhibit different limitations as follows:}
\begin{itemize}
\item PRM+D and IRN fail to detect integral instances with inferior detected boundaries (\eg, rows 1, 13).
\item C2SNet and NLDF tend to recognize texture boundaries, causing fragmented instances (\eg, rows 10, 12, 14).
\item DeepMask and S4Net suffer from the over-detection problem, as they fail to distinguish instance proposals belonging to the same instance (\eg, rows 2, 3, 13).
\item MAP is a bounding-box based method. It fails to get clear instance boundaries even post-processed by CRF (\eg, rows 1, 2, 7).
\end{itemize}

Overall, our method outperforms all these baselines, as a result of the centroid-based subitizing loss, \tx{the carefully designed BE and CA modules, and the progress training scheme.}

\subsection{Internal Analysis and Discussions}
\subsubsection{\txx{Ablation Study of Network Design}}
\kk{We begin by investigating the effectiveness of the proposed network design, including the proposed Boundary Enhancement module, Cross-layer Attention module, Progressive Training Scheme, and $\mathcal{L}_{\mathcal{SU}}$ loss. Table~\ref{tab:ab1} shows the results.}
%
We can see that the SID performance would drop if we remove any of the components from the network.
This shows that these components can help boost the performances of the saliency, centroid and boundary detection sub-tasks, which play a vital role in detecting salient instances. Figures~\ref{fig:wsu},~\ref{fig:wbe}, and~\ref{fig:da} provide additional visual comparisons to demonstrate the effectiveness of these components.

\begin{table}[h]
\caption{\tx{Ablation study of network design.}}
\vspace{-3mm}
\begin{center}
\begin{tabular}{c c c}
\toprule[1.5pt]
\textbf{method} & \textbf{mAP@0.5$\uparrow$} & \textbf{mAP@0.7$\uparrow$} \\
\midrule[1.0pt]
Ours (w/o CA, BE, PTS, $\mathcal{L}_{\mathcal{SU}}$) & 57.1\% & 37.4\% \\
Ours (w/o CA) & 64.3\% & 48.4\% \\
Ours (w/o BE) & 65.2\% & 48.9\% \\
Ours (w/o $\mathcal{L}_{\mathcal{SU}}$) & 62.1\% & 46.9\% \\
Ours (w/o PTS) & 63.9.\% & 47.2\% \\
Ours  & \textbf{68.3\%} & \textbf{51.7\%} \\
\bottomrule[1.5pt]
\end{tabular}
\end{center}
\label{tab:ab1}
\vspace{-3mm}
\end{table}

\subsubsection{\txx{Evaluation of the CA Module}}
\txx{We then investigate our CA module on its design choices and intermediate feature visualization.}

\txx{\textbf{Design Choices:} we examine our CA module against its variants, as reported in Table~\ref{tab:ab3}.}
%
First, we study the benefits of the \txx{CFM} and MA units. Row 1 shows that removing the \txx{CFM} unit leads to a performance drop, due to the reduction in communication enhancement among the pyramid feature layers. Row 2 also shows a performance drop without the MA unit, as we are not able to learn long-range dependencies.
Second, we study the connection styles (parallel and cascade) of the attention mechanisms in the MA unit. Rows 3, 4 and 5 show that parallel connection of the spatial- and channel-wise attentions performs better than the cascade one. This may be because cascade connection may lose the learned context information of the former attention mechanism.
\vspace{-2mm}

\begin{table}[!h]
\caption{\tx{Evaluation of different designs of the CA module. s$\rightarrow$c represents using spatial-wise attention before channel-wise attention, while c$\rightarrow$s represents the reverse connection. The best performance among different designs is marked in \textbf{bold}.}}
\vspace{-5mm}
\begin{center}
\begin{tabular}{c c c}
\toprule[1.5pt]
\textbf{method} & \textbf{mAP@0.5$\uparrow$} & \textbf{mAP@0.7$\uparrow$} \\
\midrule[1pt]
w/o CL, w/ parallel MA               & 66.1\% & 49.5\% \\
w/ CL, w/o parallel MA      & 66.8\%  & 50.2\%  \\
w/ CL, w/ cascade (s$\rightarrow$c) MA      & 67.4\%  & 51.4\%  \\
w/ CL, w/ cascade (c$\rightarrow$s) MA      & 67.2\%  & 51.0\%  \\
Ours (w/ CL, w/ parallel MA)   & {\bf 68.3\%}  & {\bf 51.7\%}  \\
\bottomrule[1.5pt]
\end{tabular}
\end{center}
\label{tab:ab3}
\vspace{-5mm}
\end{table}

\begin{figure*}[!h]
\centering
\includegraphics[width=.15\textwidth]{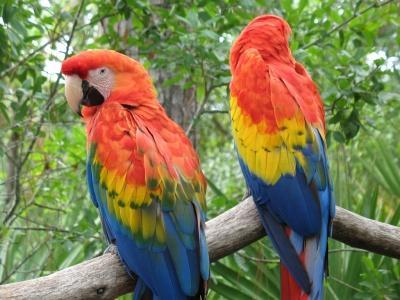}
\includegraphics[width=.15\textwidth]{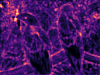}
\includegraphics[width=.15\textwidth]{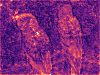}
\includegraphics[width=.15\textwidth]{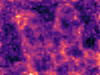}
\includegraphics[width=.15\textwidth]{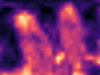}
\includegraphics[width=.15\textwidth]{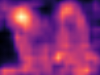}
\vspace{0.1cm}
\\
\centering
\begin{minipage}[t]{0.15\textwidth}
\centering
{input: $\mathcal{I}$}
\end{minipage}
\begin{minipage}[t]{0.15\textwidth}
\centering
{\textbf{$f_{1}$}}
\end{minipage}
\begin{minipage}[t]{0.15\textwidth}
\centering
{\textbf{$f_{2}$}}
\end{minipage}
\begin{minipage}[t]{0.15\textwidth}
\centering
{\textbf{$f_{3}$}}
\end{minipage}
\begin{minipage}[t]{0.15\textwidth}
\centering
{\textbf{$f_{4}$}}
\end{minipage}
\begin{minipage}[t]{0.15\textwidth}
\centering
{\textbf{$f_{5}$}}
\end{minipage}
\vspace{0.1cm}
\\
\centering
\includegraphics[width=.15\textwidth]{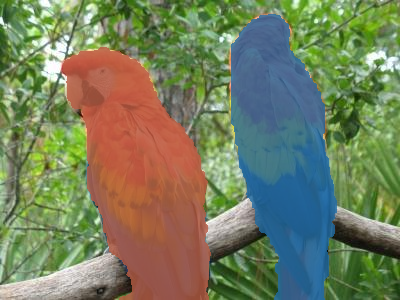}
\includegraphics[width=.15\textwidth]{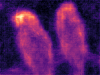}
\includegraphics[width=.15\textwidth]{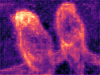}
\includegraphics[width=.15\textwidth]{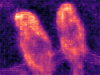}
\includegraphics[width=.15\textwidth]{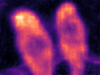}
\includegraphics[width=.15\textwidth]{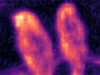}
\\
\centering
\begin{minipage}[t]{0.15\textwidth}
\centering
{inferred $\overline{\mathcal{SI}}^{*}$}
\end{minipage}
\begin{minipage}[t]{0.15\textwidth}
\centering
{\textbf{${f_1}^{*}$}}
\end{minipage}
\begin{minipage}[t]{0.15\textwidth}
\centering
\textbf{${f_2}^{*}$}
\end{minipage}
\begin{minipage}[t]{0.15\textwidth}
\centering
{\textbf{${f_3}^{*}$}}
\end{minipage}
\begin{minipage}[t]{0.15\textwidth}
\centering
{\textbf{${f_4}^{*}$}}
\end{minipage}
\begin{minipage}[t]{0.15\textwidth}
\centering
{\textbf{${f_5}^{*}$}}
\end{minipage}
\vspace{0.1cm}
\\
\centering
\includegraphics[width=.15\textwidth]{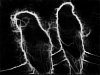}
\includegraphics[width=.15\textwidth]{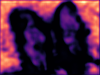}
\includegraphics[width=.15\textwidth]{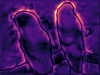}
\includegraphics[width=.15\textwidth]{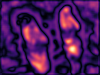}
\includegraphics[width=.15\textwidth]{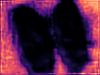}
\includegraphics[width=.15\textwidth]{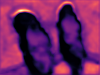}
\\
\centering
\begin{minipage}[t]{0.15\textwidth}
\centering
{predicted $\mathcal{B}$}
\end{minipage}
\begin{minipage}[t]{0.15\textwidth}
\centering
{${f_{1\Rightarrow\mathcal{B}}^{*}}$}
\end{minipage}
\begin{minipage}[t]{0.15\textwidth}
\centering
{${f_{2\Rightarrow\mathcal{B}}^{*}}$}
\end{minipage}
\begin{minipage}[t]{0.15\textwidth}
\centering
{${f_{3\Rightarrow\mathcal{B}}^{*}}$}
\end{minipage}
\begin{minipage}[t]{0.15\textwidth}
\centering
{${f_{4\Rightarrow\mathcal{B}}^{*}}$}
\end{minipage}
\begin{minipage}[t]{0.15\textwidth}
\centering
{${f_{5\Rightarrow\mathcal{B}}^{*}}$}
\end{minipage}
\vspace{0.1cm}
\\
\centering
\includegraphics[width=.15\textwidth]{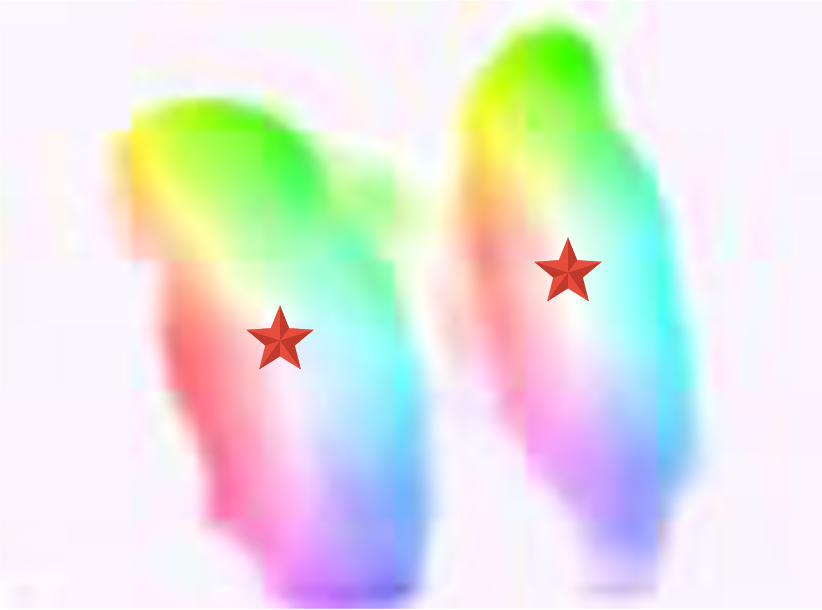}
\includegraphics[width=.15\textwidth]{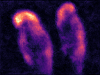}
\includegraphics[width=.15\textwidth]{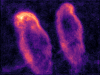}
\includegraphics[width=.15\textwidth]{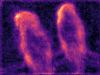}
\includegraphics[width=.15\textwidth]{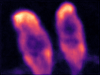}
\includegraphics[width=.15\textwidth]{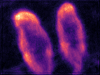}
\vspace{0.1cm}
\\
\centering
\begin{minipage}[t]{0.15\textwidth}
\centering
{{predicted $\mathcal{V}$}}
\end{minipage}
\begin{minipage}[t]{0.15\textwidth}
\centering
{${f_{1\Rightarrow\mathcal{V}}^{*}}$}
\end{minipage}
\begin{minipage}[t]{0.15\textwidth}
\centering
{${f_{2\Rightarrow\mathcal{V}}^{*}}$}
\end{minipage}
\begin{minipage}[t]{0.15\textwidth}
\centering
{${f_{3\Rightarrow\mathcal{V}}^{*}}$}
\end{minipage}
\begin{minipage}[t]{0.15\textwidth}
\centering
{${f_{4\Rightarrow\mathcal{V}}^{*}}$}
\end{minipage}
\begin{minipage}[t]{0.15\textwidth}
\centering
{${f_{5\Rightarrow\mathcal{V}}^{*}}$}
\end{minipage}
\vspace{0.1cm}
\caption{\txx{Visualization of the multi-level intermediate features learned by our CA module.}}
\label{fig:ca_vis}
\end{figure*}

\txx{\textbf{Feature Visualization:} we visualize multi-level intermediate features learned by our CA module in Figure~\ref{fig:ca_vis}. Given multi-level backbone features $f_1~\sim f_5$ ($1^{st}~row$), the CFM unit in the CA module first generates multi-level mixed context features $f_1^*~\sim f_5^*$ ($2^{nd}~row$), which are then used for learning boundary features $f_{1\Rightarrow\mathcal{B}}^{*}~\sim f_{5\Rightarrow\mathcal{B}}^{*}$ ($3^{rd}~row$) and centroid features $f_{1\Rightarrow\mathcal{V}}^{*}~\sim f_{5 \Rightarrow	\mathcal{V}}^{*}$ ($4^{th}~row$), respectively.
First, we can see that our CFM unit is able to highlight the salient objects via aggregating multi-level backbone features, as shown in row 2.
Second, the boundary-aware feature maps have high responses in different regions as shown in row 3, which suggests that determining the instance boundaries also require multi-level features.
Third, in row 4, the visualization of centroid-aware features generally corresponds to the centroid map $\mathcal{V}$, where the instance boundaries are generally highlighted, and pixel values of the centroid locations are close to zero.
Overall, our proposed CA module is able to help adapt the backbone features into different task-specific features.
} 

\begin{figure*}[!h]
\centering
\includegraphics[width=.105\textwidth]{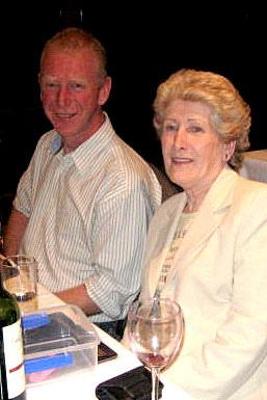}
\includegraphics[width=.105\textwidth]{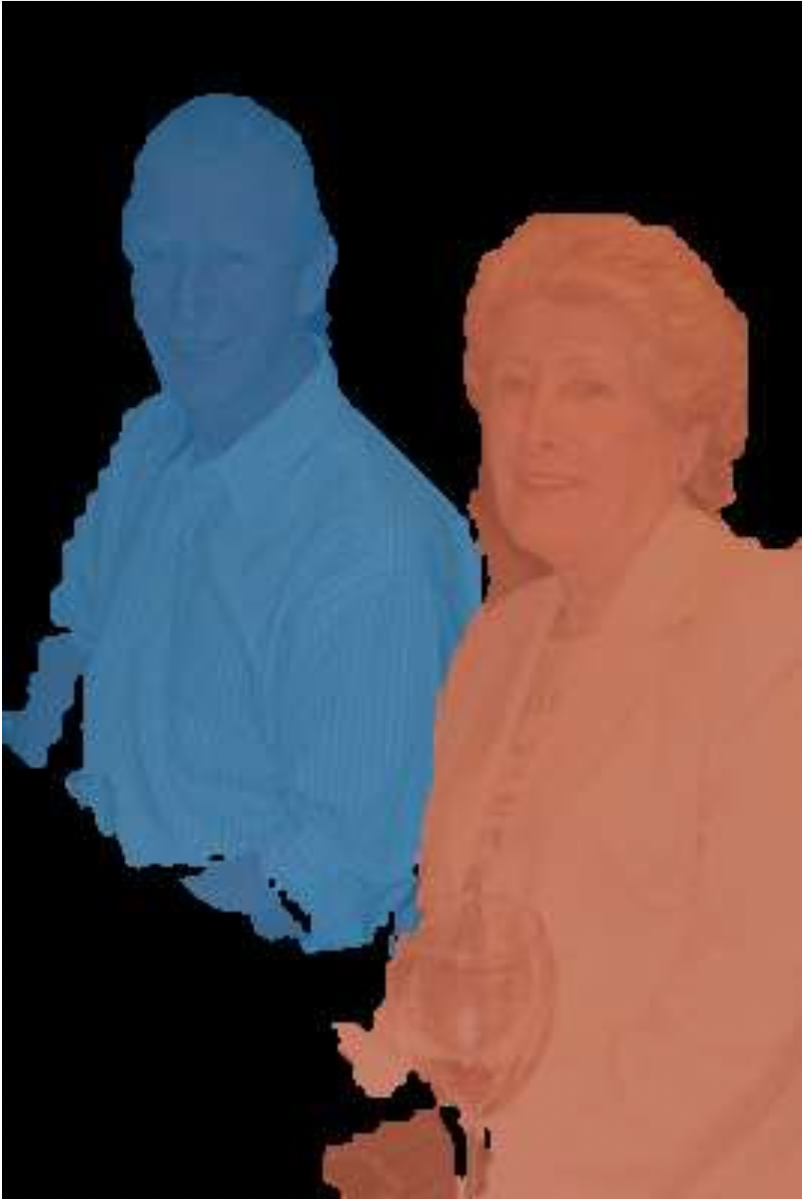}
\includegraphics[width=.105\textwidth]{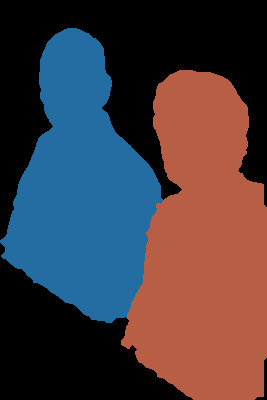}
\vspace{0.1cm}
\centering
\includegraphics[width=.105\textwidth]{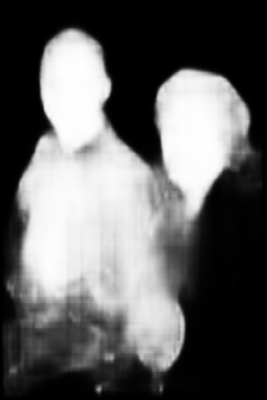}
\includegraphics[width=.105\textwidth]{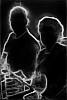}
\includegraphics[width=.105\textwidth]{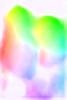}
\centering
\includegraphics[width=.105\textwidth]{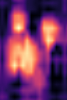}
\includegraphics[width=.105\textwidth]{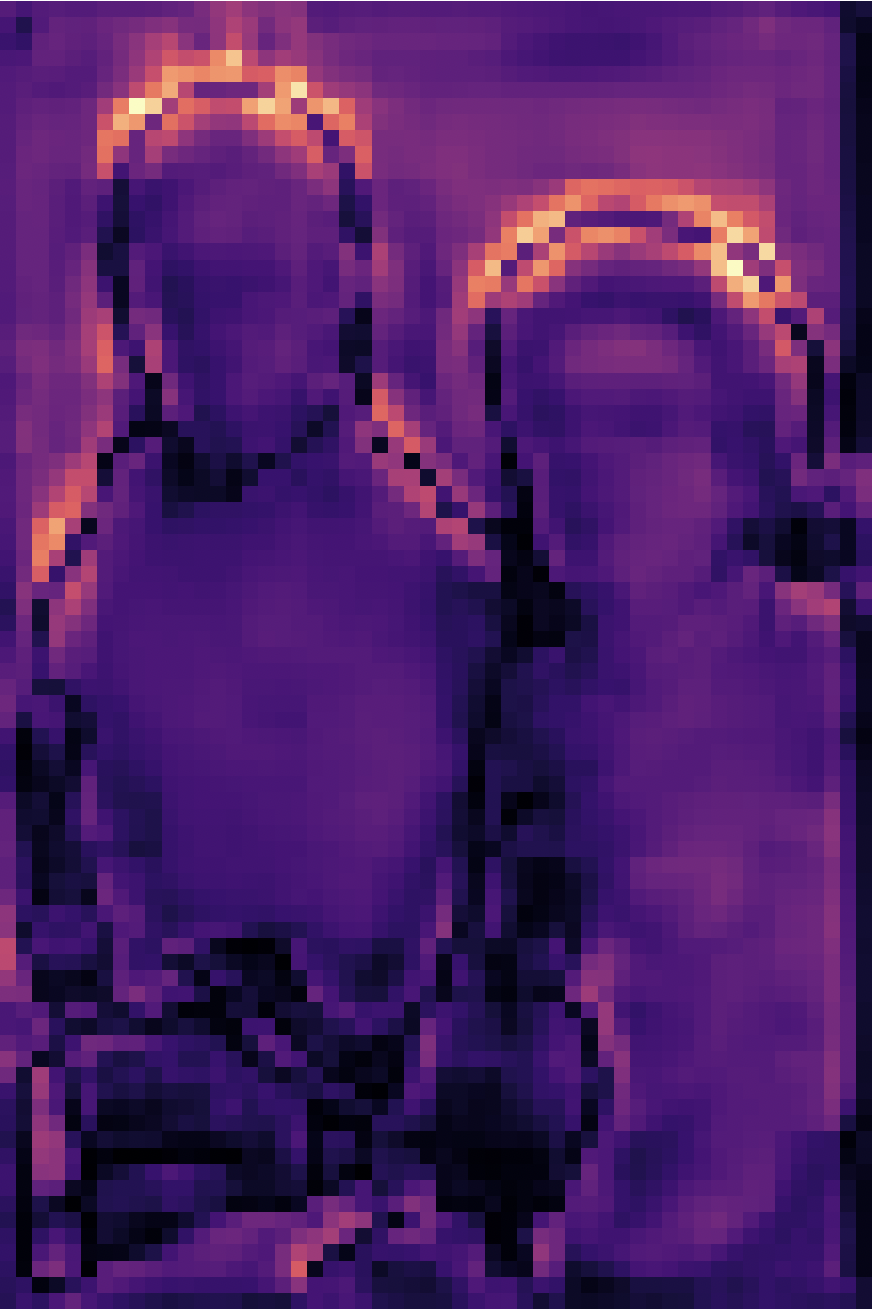}
\includegraphics[width=.105\textwidth]{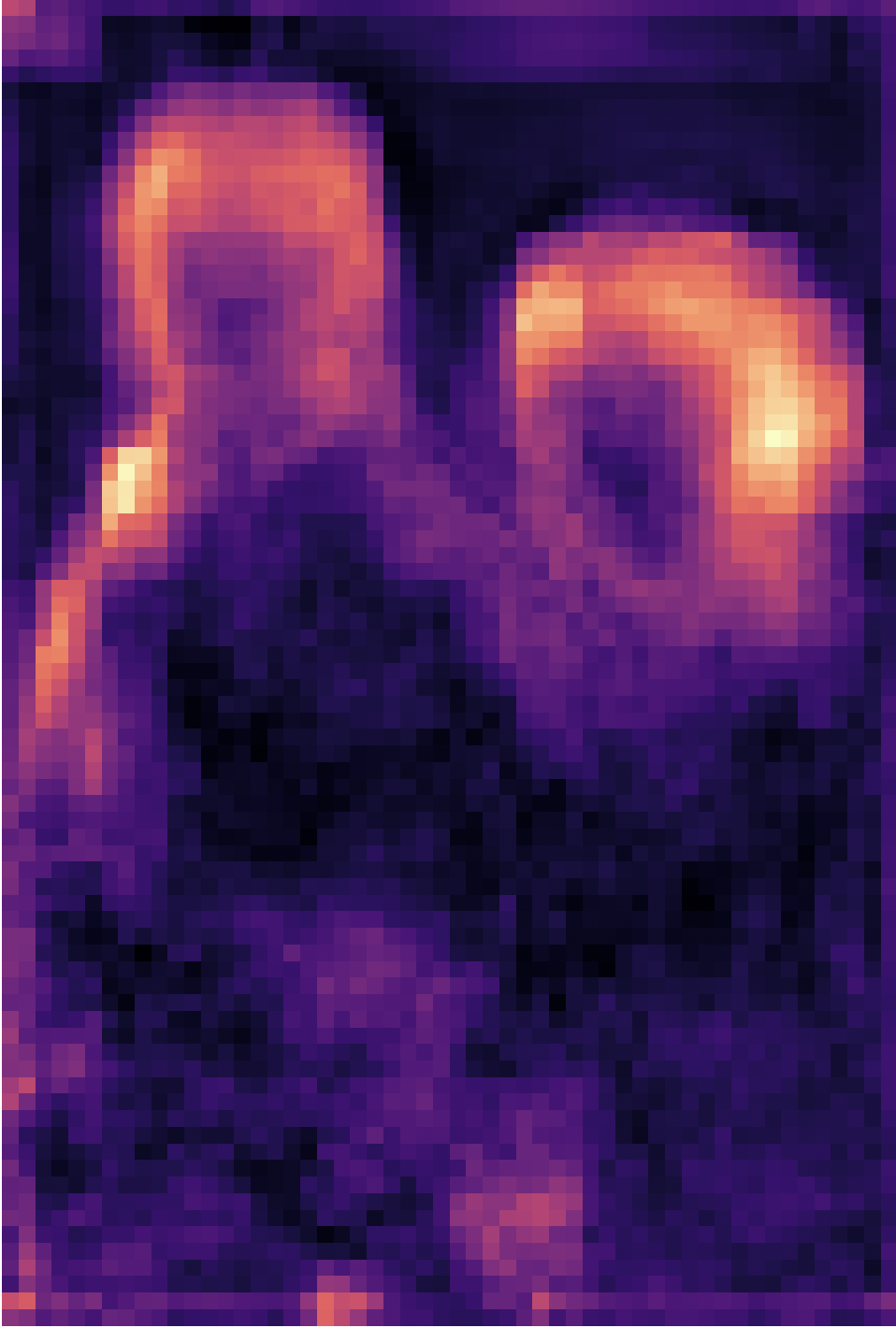}
\vspace{0.1cm}
\\
\centering
\begin{minipage}[t]{0.105\textwidth}
\centering
{input: $\mathcal{I}$}
\end{minipage}
\begin{minipage}[t]{0.105\textwidth}
\centering
{inferred $\overline{\mathcal{SI}}^{*}$}
\end{minipage}
\begin{minipage}[t]{0.105\textwidth}
\centering
{ground truth}
\end{minipage}
\centering
\begin{minipage}[t]{0.105\textwidth}
\centering
{predicted $\mathcal{S}$}
\end{minipage}
\begin{minipage}[t]{0.105\textwidth}
\centering
{predicted $\mathcal{B}$}
\end{minipage}
\begin{minipage}[t]{0.105\textwidth}
\centering
{predicted $\mathcal{V}$}
\end{minipage}
\begin{minipage}[t]{0.105\textwidth}
\centering
{CAM}
\end{minipage}
\begin{minipage}[t]{0.105\textwidth}
\centering
{CA: ${f_{2\Rightarrow\mathcal{B}}^{*}}$}
\end{minipage}
\begin{minipage}[t]{0.105\textwidth}
\centering
{CA: ${f_{3\Rightarrow\mathcal{V}}^{*}}$}
\end{minipage}
\vspace{0.1cm}
\caption{\txx{Visual comparison between CAM and CA features. As shown in column 7, CAM itself cannot delineate the boundaries between two persons and locate their centroids. Hence, we only use CAM in the saliency detection branch. Our CA module successfully learns this information for complementing CAM.}}
\label{fig:ca_vs_cam}
\end{figure*}

\txx{
In addition, our CA module differs from CAM in two aspects. 
First, CAM is conditioned on the class label input, but our CA module learns class-agnostic attentions from pseudo labels. 
Second, CAM is unable to delineate clear boundary and instance information, while our CA module can learn this information to complement the CAM for detecting salient instances, as shown in Figure~\ref{fig:ca_vs_cam}.
}

\subsubsection{\txx{Evaluation of the BE Module}}
\txx{
We conduct ablation studies to investigate the effect of the Canny filter in the Boundary Enhancement (BE) module. We compare our BE module to two ablated versions: removing the Canny filter from the BE module (denoted as BE Module w/o Canny), and using Canny filter only (denoted as Canny Only). Results are shown in Table~\ref{tab:ab_be}. We can see that our method outperforms both ablated versions. The Canny filter is used to enrich the high-level boundary features with low-level edge information. Without the Canny filter, the BE module may not detect small boundaries accurately. However, relying only on the Canny filter cannot obtain high-level boundary information, which typically leads to over-segmentation of instances.
}

\begin{table}[h]
\caption{\txx{Evaluation on the Canny filter. Best performances are marked in \textbf{bold}.}}
\vspace{-5mm}
\begin{center}
\begin{tabular}{c c c}
\toprule[1.5pt]
\textbf{method} & \textbf{mAP@0.5$\uparrow$} & \textbf{mAP@0.7$\uparrow$} \\
\midrule[1pt]
{BE Module w/o Canny}  & 65.0\% & 49.3\% \\
\midrule[0.8pt]
{Canny Only }  & 63.7\% & 48.0\% \\
\midrule[0.8pt]
{Ours}   & {\bf 68.3\%}  & {\bf 51.7\%}  \\
\bottomrule[1.5pt]
\end{tabular}
\end{center}
\label{tab:ab_be}
\vspace{-8mm}
\end{table}

\subsubsection{\txx{Evaluation of Parameter settings for the Canny Filter}}
\txx{
We empirically set the thresholds (\ie, $\theta_{up}$ and  $\theta_{low}$ for controlling the connectivity and density of the detected edges) in the Canny operator to be automatically determined by the channel median of the gray-scale image. We find that this works well in our experiments.
}

\txx{
To further investigate how these two thresholds affect the performance, we provide both qualitative and quantitative comparisons between our threshold choice with two manual choices described below:
}
\begin{itemize}
    \item \txx{Large range: we manually set $\theta_{low}$ and $\theta_{up}$ to 30 and 200, respectively, so that the Canny operator is sensitive to textures and can detect more edges.}
    \item \txx{Small range with large values: we manually set $\theta_{low}$ and  $\theta_{up}$ to 230 and 260, respectively, so that only structural edges of objects can be detected.}
\end{itemize}


\begin{table}[!t]
\caption{\txx{Evaluation on different parameter settings for the Canny operator in the BE module. Best performances are marked in \textbf{bold}.}}
\vspace{-5mm}
\begin{center}
\begin{tabular}{c c c}
\toprule[1.5pt]
\textbf{settings of $\theta_{low}$ and $\theta_{up}$} & \textbf{mAP@0.5$\uparrow$} & \textbf{mAP@0.7$\uparrow$} \\
\midrule[1pt]
{$\theta_{low}=30$, and $\theta_{up}=200$ }               & 67.4\% & 50.8\% \\ 
\midrule[0.8pt]
{$\theta_{low}=230$, and $\theta_{up}=260$ }      & 66.9\%  & 50.2\%  \\ 
\midrule[0.8pt]
{\textbf{Ours}}   & {\bf 68.3\%}  & {\bf 51.7\%}  \\ 
\bottomrule[1.5pt]
\end{tabular}
\end{center}
\label{tab:ab_theta}
\end{table}

\begin{figure}[!h]
\centering
\includegraphics[width=.115\textwidth]{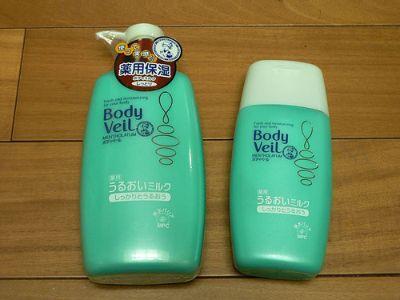}
\includegraphics[width=.115\textwidth]{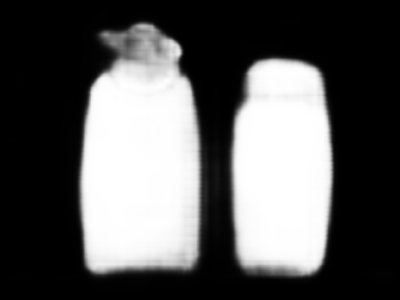}
\includegraphics[width=.115\textwidth]{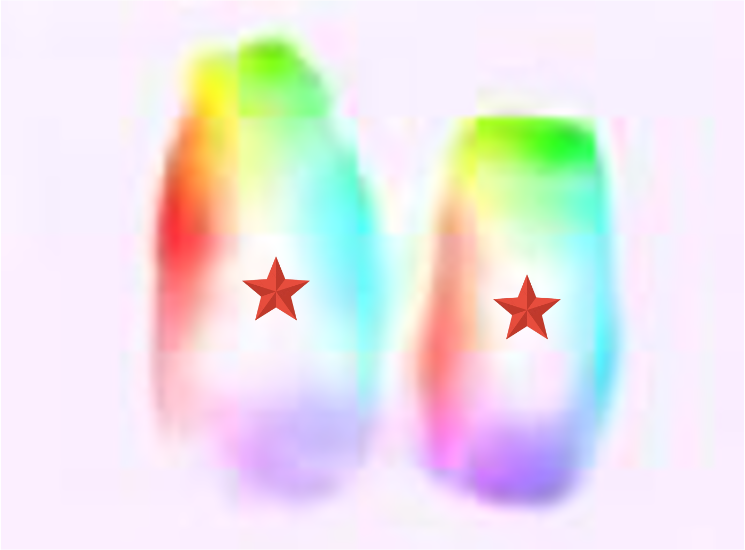}
\includegraphics[width=.115\textwidth]{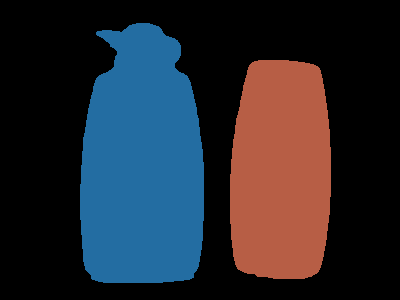}
\vspace{0.1cm}
\\
\centering
\begin{minipage}[t]{0.115\textwidth}
\centering
{\textbf{(a) input: $\mathcal{I}$}}
\end{minipage}
\begin{minipage}[t]{0.115\textwidth}
\centering
{\textbf{(b) predicted $\mathcal{S}$}}
\end{minipage}
\begin{minipage}[t]{0.115\textwidth}
\centering
{\textbf{(c) predicted $\mathcal{V}$}}
\end{minipage}
\begin{minipage}[t]{0.115\textwidth}
\centering
{\textbf{(d) GT}}
\end{minipage}
\vspace{0.1cm}
\\
\centering
\includegraphics[width=.153\textwidth]{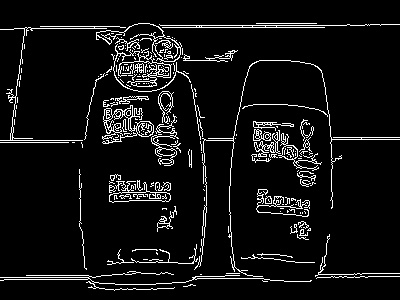}
\includegraphics[width=.153\textwidth]{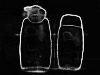}
\includegraphics[width=.153\textwidth]{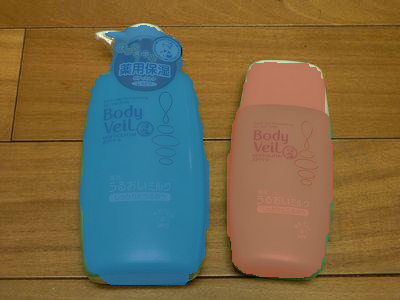}
\vspace{0.1cm}
\\
\centering
\begin{minipage}[t]{0.153\textwidth}
\centering
{\textbf{(e1) canny edge 1 \\ (\tiny{$\theta_{low}=30$ \& $\theta_{up}=200$}})}
\end{minipage}
\begin{minipage}[t]{0.153\textwidth}
\centering
{\textbf{(f1) predicted $\mathcal{B}$ \\ (using canny edge 1) }}
\end{minipage}
\begin{minipage}[t]{0.153\textwidth}
\centering
{\textbf{(g1) {inferred $\overline{\mathcal{SI}}^{*}$} \\ (using canny edge 1)}}
\end{minipage}
\vspace{0.1cm}
\\
\centering
\includegraphics[width=.153\textwidth]{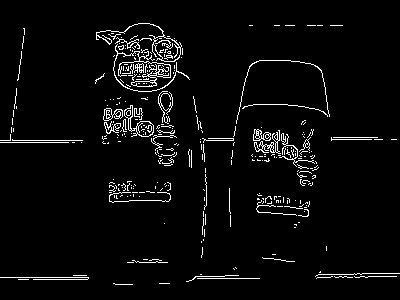}
\includegraphics[width=.153\textwidth]{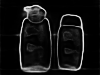}
\includegraphics[width=.153\textwidth]{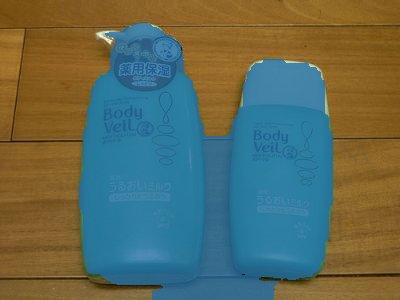}
\vspace{0.1cm}
\\
\centering
\begin{minipage}[t]{0.153\textwidth}
\centering
{\textbf{(e2) canny edge 2 \\ \tiny{($\theta_{low}=230$ \& $\theta_{up}=260$)}}}
\end{minipage}
\begin{minipage}[t]{0.153\textwidth}
\centering
{\textbf{(f2) predicted $\mathcal{B}$ \\ (using canny edge 2) }}
\end{minipage}
\begin{minipage}[t]{0.153\textwidth}
\centering
{\textbf{(g2) {inferred $\overline{\mathcal{SI}}^{*}$} \\ (using canny edge 2)}}
\end{minipage}
\vspace{0.1cm}
\\
\centering
\includegraphics[width=.153\textwidth]{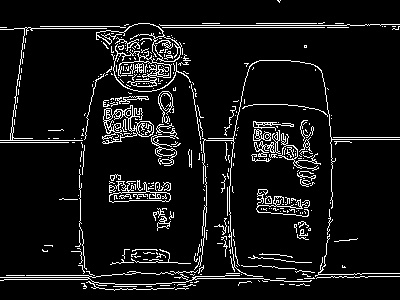}
\includegraphics[width=.153\textwidth]{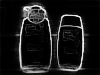}
\includegraphics[width=.153\textwidth]{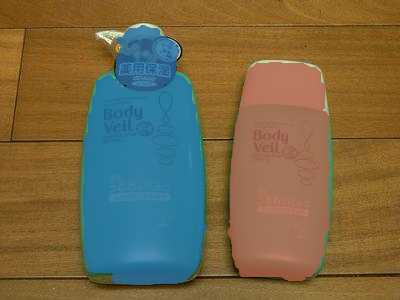}
\vspace{0.1cm}
\\
\centering
\begin{minipage}[t]{0.153\textwidth}
\centering
{\textbf{(e3) canny edge 3 \\ (\scriptsize{automatic generated \\ $\theta_{low}$ \& $\theta_{up}$}})}
\end{minipage}
\begin{minipage}[t]{0.153\textwidth}
\centering
{\textbf{(f3) predicted $\mathcal{B}$ \\ (using canny edge 3) }}
\end{minipage}
\begin{minipage}[t]{0.153\textwidth}
\centering
{\textbf{(g3) {inferred $\overline{\mathcal{SI}}^{*}$} \\ (using canny edge 3)}}
\end{minipage}
\vspace{0.1cm}
\\
\caption{\txx{Visual comparison between results using different parameters of the Canny operator.}}
\label{fig:canny_params_1}
\vspace{-3mm}
\end{figure}

\begin{figure}[!h]
\centering
\includegraphics[width=.115\textwidth]{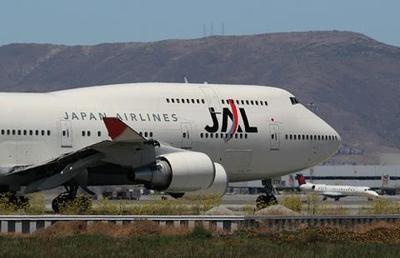}
\includegraphics[width=.115\textwidth]{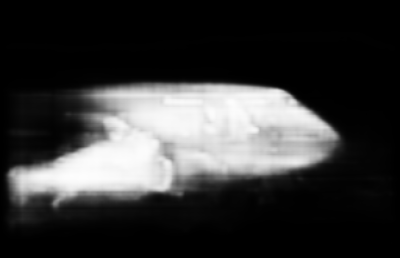}
\includegraphics[width=.115\textwidth]{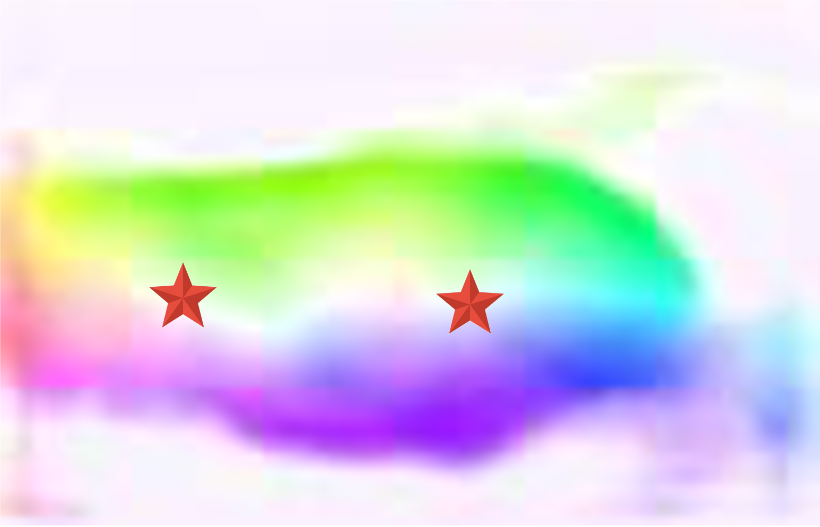}
\includegraphics[width=.115\textwidth]{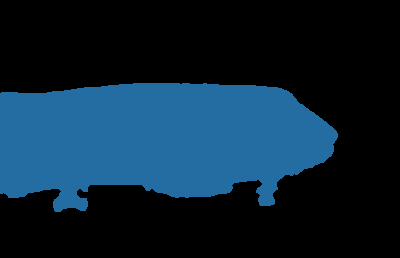}
\vspace{0.1cm}
\\
\centering
\begin{minipage}[t]{0.115\textwidth}
\centering
{\textbf{(a) input: $\mathcal{I}$}}
\end{minipage}
\begin{minipage}[t]{0.115\textwidth}
\centering
{\textbf{(b) predicted $\mathcal{S}$}}
\end{minipage}
\begin{minipage}[t]{0.115\textwidth}
\centering
{\textbf{(c) predicted $\mathcal{V}$}}
\end{minipage}
\begin{minipage}[t]{0.115\textwidth}
\centering
{\textbf{(d) GT}}
\end{minipage}
\vspace{0.1cm}
\\
\centering
\includegraphics[width=.153\textwidth]{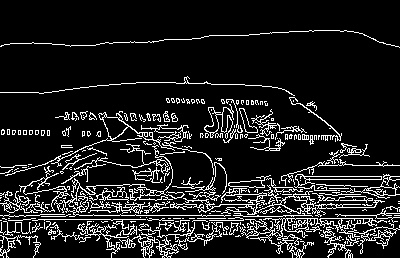}
\includegraphics[width=.153\textwidth]{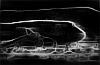}
\includegraphics[width=.153\textwidth]{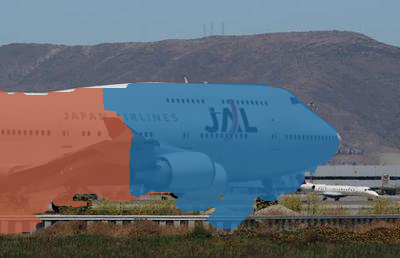}
\vspace{0.1cm}
\\
\centering
\begin{minipage}[t]{0.153\textwidth}
\centering
{\textbf{(e1) canny edge 1 \\ (\tiny{$\theta_{low}=30$ \& $\theta_{up}=200$}})}
\end{minipage}
\begin{minipage}[t]{0.153\textwidth}
\centering
{\textbf{(f1) predicted $\mathcal{B}$ \\ (using canny edge 1) }}
\end{minipage}
\begin{minipage}[t]{0.153\textwidth}
\centering
{\textbf{(g1) {inferred $\overline{\mathcal{SI}}^{*}$} \\ (using canny edge 1)}}
\end{minipage}
\vspace{0.1cm}
\\
\centering
\includegraphics[width=.153\textwidth]{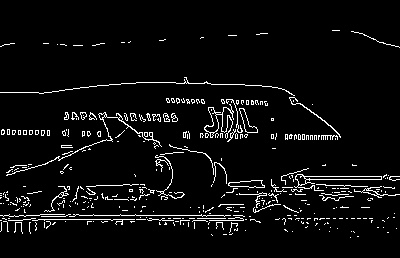}
\includegraphics[width=.153\textwidth]{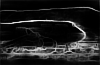}
\includegraphics[width=.153\textwidth]{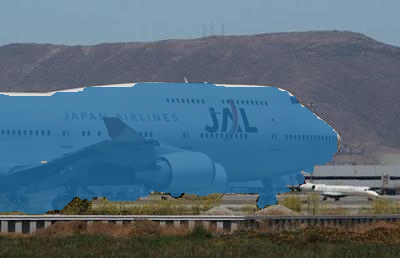}
\vspace{0.1cm}
\\
\centering
\begin{minipage}[t]{0.153\textwidth}
\centering
{\textbf{(e2) canny edge 2 \\ \tiny{($\theta_{low}=230$ \& $\theta_{up}=260$)}}}
\end{minipage}
\begin{minipage}[t]{0.153\textwidth}
\centering
{\textbf{(f2) predicted $\mathcal{B}$ \\ (using canny edge 2) }}
\end{minipage}
\begin{minipage}[t]{0.153\textwidth}
\centering
{\textbf{(g2) {inferred $\overline{\mathcal{SI}}^{*}$} \\ (using canny edge 2)}}
\end{minipage}
\vspace{0.1cm}
\\
\centering
\includegraphics[width=.153\textwidth]{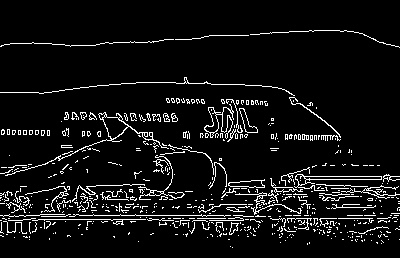}
\includegraphics[width=.153\textwidth]{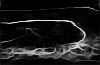}
\includegraphics[width=.153\textwidth]{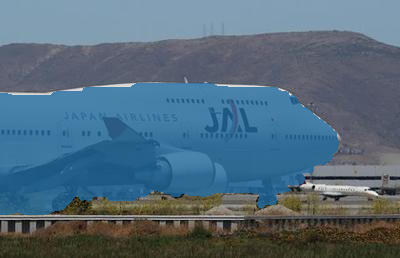}
\vspace{0.1cm}
\\
\centering
\begin{minipage}[t]{0.153\textwidth}
\centering
{\textbf{(e3) canny edge 3 \\ (\scriptsize{automatic generated $\theta_{low}$ \& $\theta_{up}$}})}
\end{minipage}
\begin{minipage}[t]{0.153\textwidth}
\centering
{\textbf{(f3) predicted $\mathcal{B}$ \\ (using canny edge 3) }}
\end{minipage}
\begin{minipage}[t]{0.153\textwidth}
\centering
{\textbf{(g3) {inferred $\overline{\mathcal{SI}}^{*}$} \\ (using canny edge 3)}}
\end{minipage}
\vspace{0.1cm}
\\
\caption{\txx{Visual comparison between results using different parameters of the Canny operator.}}
\label{fig:canny_params_2}
\end{figure}

\txx{
Table~\ref{tab:ab_theta} shows that both manual strategies would degrade the performance. Figures~\ref{fig:canny_params_1} and~\ref{fig:canny_params_2} show two scenes that these manual strategies fail. 
In row 2 of Figure~\ref{fig:canny_params_1}, if the Canny edge map provides insufficient object structure information and the learned boundaries are partially weak, our method fails to separate nearby instances. In row 1 of Figure~\ref{fig:canny_params_2}, the Canny edge contains extensive non-structure textures that affect the learned boundaries, making it difficult for the saliency values to propagate to the target boundaries from the centroid for determining the instance. In contrast, our choice successfully detects accurate instances since we can obtain high-quality boundaries, as shown in rows 1 and 3 in Figure~\ref{fig:canny_params_1}, and rows 2 and 3 in Figure~\ref{fig:canny_params_2}. This visually verifies that the Canny edge generated under our automatic setting is more stable to provide pleasant instance boundaries.
}

\subsubsection{\txx{Evaluation of PTS}}

\begin{figure}[!h]
\centering
\begin{minipage}{0.0555\textwidth}
\centering
\tiny{{\textbf{input\\image}}}
\end{minipage}
\begin{minipage}{0.0555\textwidth}
\centering
\tiny{{\textbf{Salmap\\(iter. 1)}}}
\end{minipage}
\begin{minipage}{0.0555\textwidth}
\centering
\tiny{{\textbf{Salmap\\(iter. 2)}}}
\end{minipage}
\begin{minipage}{0.0555\textwidth}
\centering
\tiny{{\textbf{Salmap\\(iter. 3)}}}
\end{minipage}
\begin{minipage}{0.0555\textwidth}
\centering
\tiny{{\textbf{Salmap\\(iter. 4)}}}
\end{minipage}
\begin{minipage}{0.0555\textwidth}
\centering
\tiny{{\textbf{Salmap\\(iter. 5)}}}
\end{minipage}
\begin{minipage}{0.0555\textwidth}
\centering
\tiny{{\textbf{Salmap\\(iter. 6)}}}
\end{minipage}
\begin{minipage}{0.0555\textwidth}
\centering
\tiny{{\textbf{ground\\truth}}}
\end{minipage}
\\
\centering
\includegraphics[width=.49\textwidth]{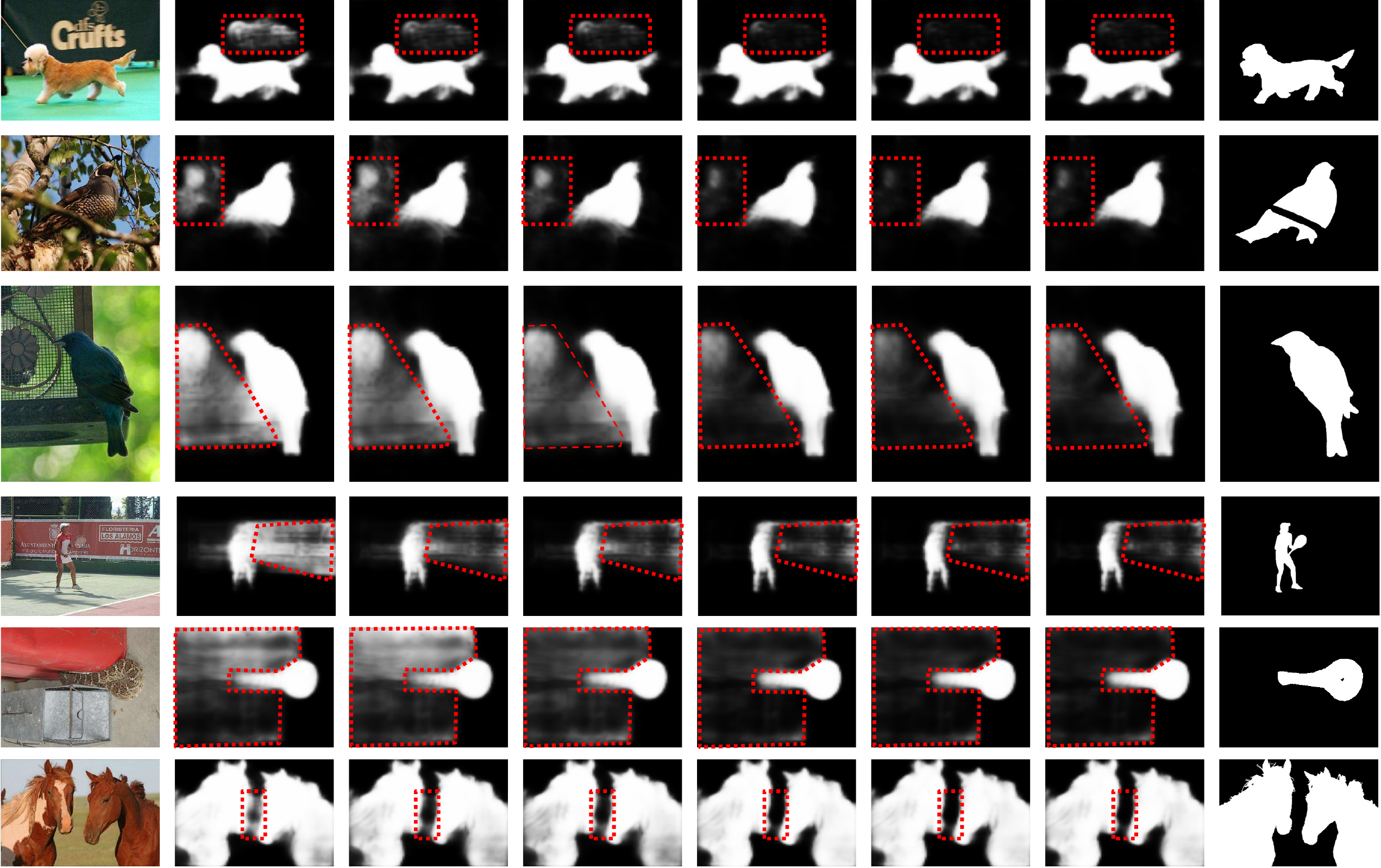}
\caption{\tx{Visualization of intermediate results over six training rounds. The dashed red regions denote background distraction noise. We can see that the noise is progressively suppressed over the iterations.}}\label{fig:PTS_res1}
\end{figure}

\begin{figure}[!h]
\centering
\begin{minipage}{0.0555\textwidth}
\centering
\tiny{{\textbf{input\\image}}}
\end{minipage}
\begin{minipage}{0.0555\textwidth}
\centering
\tiny{{\textbf{Salmap\\(iter. 1)}}}
\end{minipage}
\begin{minipage}{0.0555\textwidth}
\centering
\tiny{{\textbf{Salmap\\(iter. 2)}}}
\end{minipage}
\begin{minipage}{0.0555\textwidth}
\centering
\tiny{{\textbf{Salmap\\(iter. 3)}}}
\end{minipage}
\begin{minipage}{0.0555\textwidth}
\centering
\tiny{{\textbf{Salmap\\(iter. 4)}}}
\end{minipage}
\begin{minipage}{0.0555\textwidth}
\centering
\tiny{{\textbf{Salmap\\(iter. 5)}}}
\end{minipage}
\begin{minipage}{0.0555\textwidth}
\centering
\tiny{{\textbf{Salmap\\(iter. 6)}}}
\end{minipage}
\begin{minipage}{0.0555\textwidth}
\centering
\tiny{{\textbf{ground\\truth}}}
\end{minipage}
\\
\centering
\includegraphics[width=.49\textwidth]{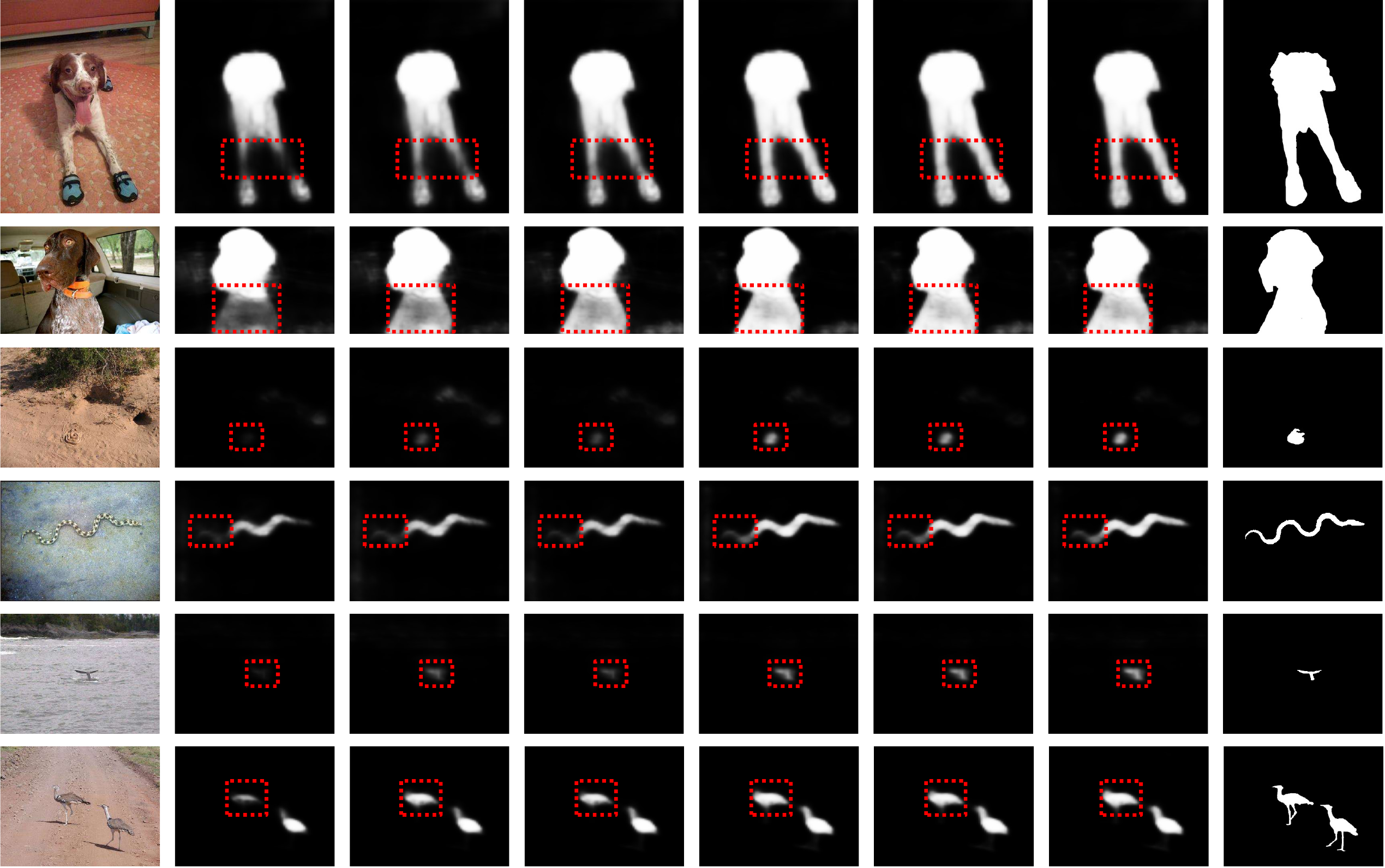}
\caption{\tx{Visualization of intermediate results over six training rounds. The dashed red regions denote missed salient parts that are progressively recovered over the iterations. We can see that the detected objects are becoming  more and more complete.}}\label{fig:PTS_res2}
\end{figure}

\tx{
We study how PTS helps improve the saliency detection performance iteration by iteration. Figure~\ref{fig:PTS_res1} and~\ref{fig:PTS_res2} show the progressively improving results over six training iterations. Intermediate results in both figures verity that our PTS does not only penalize the background distraction but also recover the integral objects.
Overall, our PTS is able to reduce noise and improve the performance in a self-supervised manner.
}

\subsubsection{\tx{Evaluation of the Saliency Detection Branch}}
\kk{Since our WSID-Net relies on the performance of the saliency detection branch in detecting salient objects, we are particularly interested in the question of to what extent that the quality of the saliency object detection maps may affect the SID performance.
%
%
To answer this question, we replace the outputs of our weakly-supervised salient object detection branch with five state-of-the-art full-supervised SOD methods (\ie, DSS~\cite{wang2018detect}, PiCANet~\cite{liu2018picanet}, EGNet~\cite{zhao2019egnet}, ITSD~\cite{zhou2020interactive}, and SCRN~\cite{wu2019stacked}), as well as the ground truth saliency maps, to generate the salient instance maps. Results are reported in Table~\ref{tab:ab2}.
%
%
\tx{
We can see that the performance generally increases when inferring salient instance masks using the fully-supervised SOD results. This is because the fully-supervised methods are more robust to background distractions and able to delineate full object masks.}
However, we can still observe that the performance would not be saturated even if we feed the ground truth saliency maps to generate the SID maps. This is because the instance boundaries are still very difficult to detect, especially when these salient instances overlap each other. This suggests that developing an effective method for detecting salient instance boundaries in a weakly-supervised setting would be a promising solution.
}

\begin{table}[!t]
\caption{\kk{Investigation on how the SID performance is affected by the quality of the SOD maps. We show the performances by replacing the saliency maps (denoted Salmap) predicted by our saliency detection branch with saliency maps computed by different fully-supervised state-of-the-art SOD methods.}}
\vspace{-3mm}
\begin{center}
\begin{tabular}{c c c}
\toprule[1.5pt]
\textbf{method} & \textbf{mAP@0.5$\uparrow$} & \textbf{mAP@0.7$\uparrow$} \\
\midrule[1.0pt]
Salmap $\rightarrow$ GT                & 72.1\% & 58.3\% \\
Salmap $\rightarrow$ DSS~\cite{wang2018detect}       & 67.2\%  & 54.3\%  \\
Salmap $\rightarrow$ PiCANet~\cite{liu2018picanet}   & 67.9\%  & 53.8\%  \\
Salmap $\rightarrow$ EGNet \cite{zhao2019egnet}      & 69.3\%  & 54.9\%  \\
Salmap $\rightarrow$ ITSD~\cite{zhou2020interactive} & 70.0\%  & 56.4\%  \\
Salmap $\rightarrow$ SCRN \cite{wu2019stacked}       & 69.2\% & 55.9\% \\
Ours                                    & 68.3\% & 51.7\% \\
\bottomrule[1.5pt]
\end{tabular}
\end{center}
\label{tab:ab2}
\vspace{-3mm}
\end{table}

\subsubsection{\txx{Evaluation of CRF in the Saliency Branch}}
\begin{figure}[h]
\centering
\begin{minipage}[t]{0.1\textwidth}
\centering
{\textbf{input image}}
\end{minipage}
\begin{minipage}[t]{0.105\textwidth}
\centering
{\textbf{CAM}}
\end{minipage}
\begin{minipage}[t]{0.115\textwidth}
\centering
{\textbf{CRF result}}
\end{minipage}
\begin{minipage}[t]{0.1\textwidth}
\centering
{\textbf{ground truth}}
\end{minipage}
\vspace{0.1cm}
\centering
\includegraphics[width=.45\textwidth]{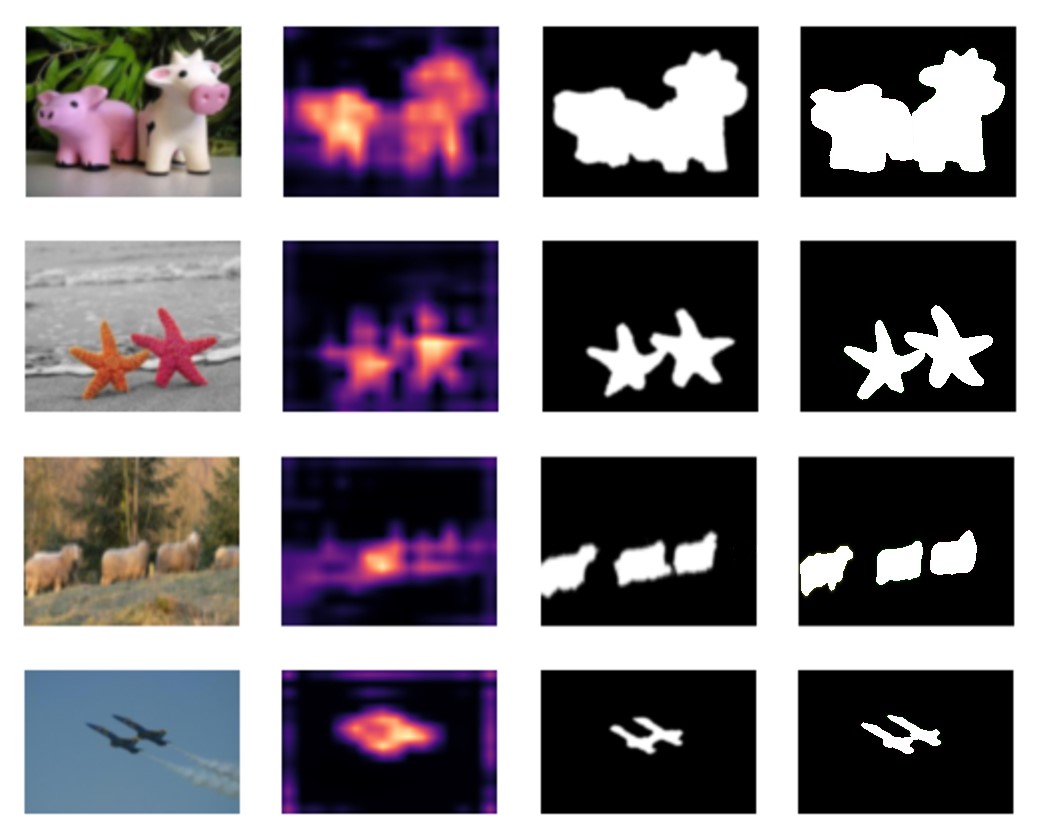}
\vspace{0.1cm}
\caption{\txx{Visualization of the effect of CRF on refining the coarse CAM.}}
\label{fig:crf_cam}
\end{figure}

\txx{
CRF is used to produce pseudo ground truth saliency maps given the coarse CAM activation maps, so that the saliency detection branch can learn more accurate boundary information. Figure~\ref{fig:crf_cam} shows that CRF helps produce more accurate pseudo ground truth saliency maps. We also provide quantitative results in Table~\ref{tab:ab_crf}, from which we can see that the performance drops without CRF refinement.
}

\begin{table}[h]
\caption{\txx{Evaluation of the effect of CRF to the saliency branch. Best performances are marked in \textbf{bold}.}}
\vspace{-5mm}
\begin{center}
\begin{tabular}{c c c}
\toprule[1.5pt]
\textbf{method} & \textbf{mAP@0.5$\uparrow$} & \textbf{mAP@0.7$\uparrow$} \\
\midrule[1pt]
{w/ CRF}  & 62.4\% & 43.1\% \\
\midrule[0.8pt]
{w/o CRF (Ours)}   & {\bf 68.3\%}  & {\bf 51.7\%}  \\
\bottomrule[1.5pt]
\end{tabular}
\end{center}
\label{tab:ab_crf}
\end{table}

\section{Conclusion and Future Work}

In this paper, we have proposed the first weakly-supervised SID method, called WSID-Net, which is trained on class and subitizing labels. Our WSID-Net learns to predict object boundaries, instance centroids, and salient regions. By using the proposed Boundary Enhancement module, \tx{Cross-layer Attention module, Progressive Training Scheme,} and centroid-based subitizing loss, our method can identify and segment salient instances effectively. Both quantitative and qualitative experiments demonstrate the effectiveness of the proposed method compared with baseline methods.

\tx{Our method does have its limitation. It may fail when the images are taken with improper exposures. Therefore, our method cannot detect salient objects/instances with low contrast to their surroundings. As a future work, we are currently exploring the use of a discriminative network of generative adversarial learning to overcome this limitation. We would also like to extend this work for videos.}

%

\begin{acknowledgements}
This work was supported in part by the National Natural Science Foundation of China under Grant 61632006, Grant 61972067, and the Innovation Technology Funding of Dalian (Project No. 2018J11CY010, 2020JJ26GX036); a General Research Fund from RGC of Hong Kong (RGC Ref.: 11205620); and a Strategic Research Grant from City University of Hong Kong (Ref.: 7005674).
\end{acknowledgements}

%
%

\bibliographystyle{spmpsci}      
\bibliography{main}   
\vspace{1in}

\section*{Author Biographies}
\vspace{-0.3in}
\begin{figure}[h]
        \begin{minipage}[t]{0.85\linewidth}
        \begin{wrapfigure}{l}{20mm}
\vspace{-25pt}
            \includegraphics[width=1.25in,height=1.2in,clip,keepaspectratio]{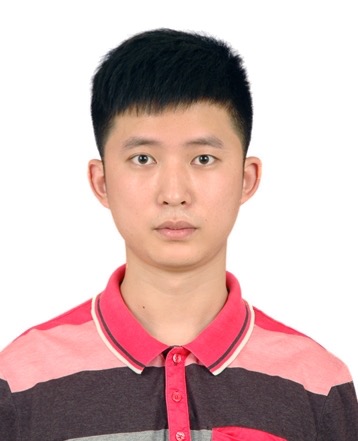}
            \end{wrapfigure}
            \textbf{Xin Tian} is a PhD student in the Department of Computer Science at Dalian University of Technology and City University of HongKong. His research interests include salient object detection and image restoration.
        \end{minipage} 
\end{figure}
\vspace{0.2in}

\begin{figure}[h]
        \begin{minipage}[t]{0.85\linewidth}
        \begin{wrapfigure}{l}{20mm}
\vspace{-25pt}
            \includegraphics[width=1.25in,height=1.2in,clip,keepaspectratio]{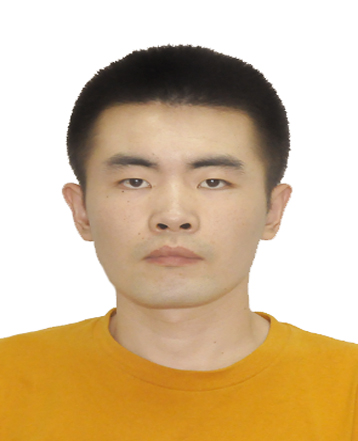}
            \end{wrapfigure}
            \textbf{Ke Xu} is currently with the Department of Computer Science at City University of Hong Kong. He obtains the dual Ph.D. degrees from Dalian University of Technology and City University of Hong Kong. His research interests include deep learning, object detection, and image enhancement and editing.
        \end{minipage} 
\end{figure}
\vspace{0.2in}

\begin{figure}[h]
        \begin{minipage}[t]{0.85\linewidth}
                \begin{wrapfigure}{l}{20mm}
            \includegraphics[width=1.25in,height=1.2in,clip,keepaspectratio]{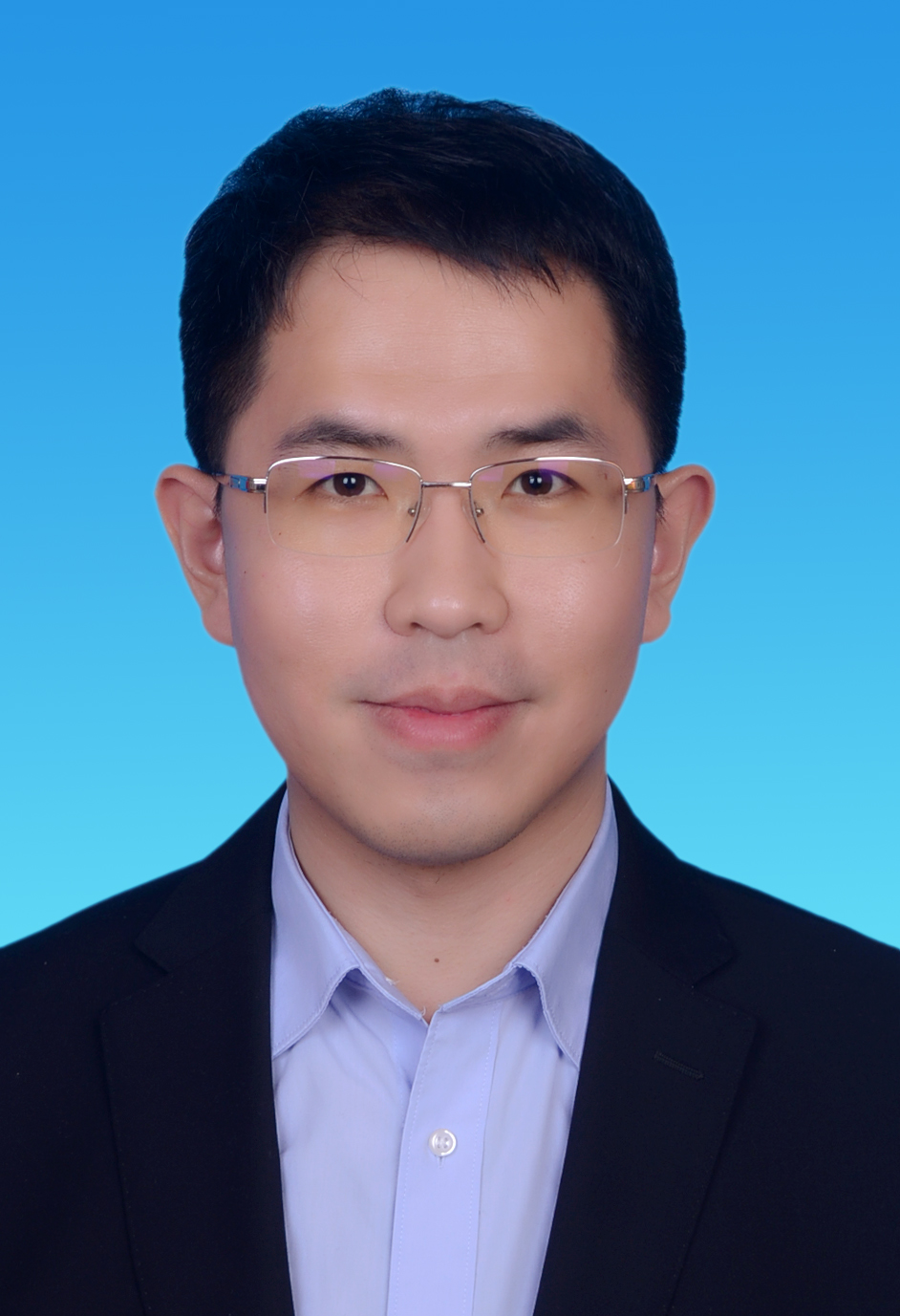}
            \end{wrapfigure}
            \textbf{Xin Yang} is a professor in the Department of Computer Science at Dalian University of Technology, China. Xin received his B.S. degree in Computer Science from Jilin University in 2007. From 2007 to June 2012, he was a joint Ph.D. student in Zhejiang University and UC Davis for Graphics, and received his Ph.D. degree in July 2012. His research interests include computer graphics and robotic vision.
        \end{minipage} 
\end{figure}
\vspace{0.2in}

\begin{figure}[h]
        \begin{minipage}[t]{0.85\linewidth}
                \begin{wrapfigure}{l}{20mm}
\vspace{-20pt}
            \includegraphics[width=1.25in,height=1.2in,clip,keepaspectratio]{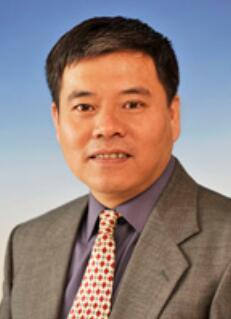}
            \end{wrapfigure}
\textbf{Baocai Yin} is a professor of computer science department at the Dalian University of Technology and the dean of Faculty of Electronic Information and Electrical Engineer. His research concentrates on digital multimedia and computer vision. He received his B.S. degree and Ph.D. degree in computer science, both from Dalian University of Technology.
        \end{minipage}        
\end{figure}
\vspace{0.2in}

\begin{figure}[h]
        \begin{minipage}[t]{0.85\linewidth}
                \begin{wrapfigure}{l}{20mm}
\vspace{-20pt}
            \includegraphics[width=1.25in,height=1.2in,clip,keepaspectratio]{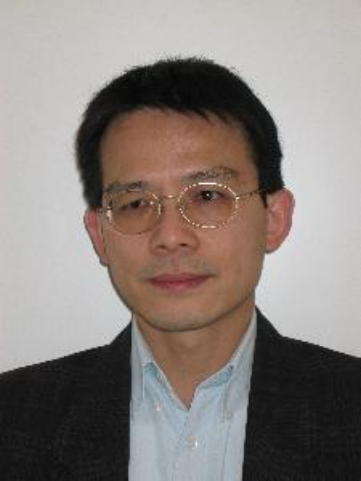}
            \end{wrapfigure}
\textbf{Rynson W.H. Lau} received his Ph.D. degree from University of Cambridge. He was on the faculty of Durham University and is now with City University of Hong Kong.

Rynson serves on the Editorial Board of the International Journal of Computer Vision (IJCV) and Computer Graphics Forum.
He has served as the Guest Editor of a number of journal special issues, including ACM Trans. on Internet Technology, IEEE Trans. on Multimedia, IEEE Trans. on Visualization and Computer Graphics, and IEEE Computer Graphics \& Applications. He has also served in the committee of a number of conferences, including Program Co-chair of ACM VRST 2004, ACM MTDL 2009, IEEE U-Media 2010, and Conference Co-chair of CASA 2005, ACM VRST 2005, ACM MDI 2009, ACM VRST 2014. Rynson's research interests include computer graphics and computer vision.
        \end{minipage}
    \end{figure}
\end{sloppypar}
\end{document}